\documentclass[sigconf]{acmart}

\AtBeginDocument{%
  }

\setcopyright{acmlicensed}
\copyrightyear{2025}
\acmYear{2025}
\acmDOI{0000.0000}
\acmConference[Preprint]{}{October 30, 2025}{Campinas/SP, Brazil}
\acmISBN{}

\usepackage{booktabs}
\usepackage{subcaption}
\usepackage{xcolor}
\usepackage{tikz}
\usepackage{listings}
\usepackage{float}

\newcommand{\ie}{\textit{i.e.}}
\newcommand{\eg}{\textit{e.g.}}
\newcommand{\etal}{\textit{et~al.}}

\newcommand*\circled[1]{\tikz[baseline=(char.base)]{
   \node[shape=circle,draw,inner sep=1pt] (char) {#1};}}

\newif\ifComments
\Commentsfalse
\ifComments
\newcommand{\del}[1]{\noindent\textcolor{gray}{{#1}}}
\newcommand{\new}[1]{\noindent\textcolor{blue}{{#1}}}

\newcommand{\guido}[1]{\noindent\textcolor{magenta}{Guido: {#1}}}
\newcommand{\lucas}[1]{\noindent\textcolor{violet}{Lucas: {#1}}}
\newcommand{\marcio}[1]{\noindent\textcolor{green}{Marcio: {#1}}}
\newcommand{\victor}[1]{\noindent\textcolor{teal}{Victor: {#1}}}
\newcommand{\gustavo}[1]{\noindent\textcolor{orange}{Gustavo: {#1}}}
\else
\newcommand{\del}[1]{}
\newcommand{\new}[1]{#1}

\newcommand{\guido}[1]{}
\newcommand{\lucas}[1]{}
\newcommand{\marcio}[1]{}
\newcommand{\victor}[1]{}
\newcommand{\gustavo}[1]{}
\fi

\newcommand{\SConvTransform}             {{SConvTransform}}

\newcommand{\SConvOp}                    {\texttt{SConvOp}}
\newcommand{\LowerToBlasOp}              {\texttt{LowerToBlasOp}}
\newcommand{\GenericOp}                  {\texttt{linalg::GenericOp}}
\newcommand{\ConvOp}                     {\texttt{linalg::Conv2DNchwFchwOp}}

\newcommand{\MlirSConv}                  {\texttt{transform.structured.sconv}}
\newcommand{\MlirSplit}                  {\texttt{transform.structured.split}}
\newcommand{\MlirGeneralize}             {\texttt{transform.structured.generalize}}
\newcommand{\MlirGeneric}                {\texttt{linalg.generic}}
\newcommand{\MlirConv}                   {\texttt{linalg.conv2d\_nchw\_fchw}}
\newcommand{\MlirConvGroup}              {\texttt{linalg.conv2d\_ngchw\_fgchw}}
\newcommand{\MlirTranspose}              {\texttt{tensor.transpose}}
\newcommand{\MlirConcat}                 {\texttt{tensor.concat}}
\newcommand{\MlirEmpty}                  {\texttt{tensor.empty}}
\newcommand{\MlirInsertSlice}            {\texttt{tensor.insert\_slice}}
\newcommand{\MlirExtractSlice}           {\texttt{tensor.extract\_slice}}
\newcommand{\MlirErf}                    {\texttt{math.erf}}
\newcommand{\MlirRsqrt}                  {\texttt{math.rsqrt}}
\newcommand{\MlirFor}                    {\texttt{scf.for}}
\newcommand{\MlirExtractStridedMetadata} {\texttt{memref.extract\_strided\_metadata}}
\newcommand{\MlirCall}                   {\texttt{func.call}}
\newcommand{\MlirContract}               {\texttt{vector.contract}}

\newcommand{\Transform}                  {{Transform}}
\newcommand{\Vector}                     {{Vector}}
\newcommand{\Memref}                     {{MemRef}}
\newcommand{\Tensor}                     {{Tensor}}
\newcommand{\Linalg}                     {{Linalg}}
\newcommand{\SCF}                        {{SCF}}
\newcommand{\Affine}                     {{Affine}}
\newcommand{\Math}                       {{Math}}

\newcommand{\MlirOpt}                    {\texttt{mlir-opt}}
\newcommand{\MlirRunner}                 {\texttt{mlir-runner}}
\newcommand{\TransformOpt}               {\texttt{transform-opt}}

\lstdefinelanguage{mlir}{
  morekeywords={module, func, tensor, ins, outs, return, linalg, generic, affine, apply, affine_map, scf, for, to, step, yield, f32, insert_slice, extract_slice},
  sensitive=true,
}

\newcommand{\sconvpar}[1]{
  \vspace{1ex}
  \noindent\textit{#1} \\
  \noindent
}

\begin{document}

\title{Using MLIR Transform to Design Sliced Convolution Algorithm}

 \author{Victor Ferrari$^{1,3}$, Marcio Pereira$^{2}$, Lucas Alvarenga$^{3}$, Gustavo Leite$^{2,3}$, Guido Araujo$^{2}$}
 \affiliation{%
    \institution{$^1$~IBM Research} 
    \institution{$^2$~Celera Systems}
    \institution{$^3$~Instituto de Computaç\~{a}o, Universidade Estadual de Campinas}
    \country{Campinas -- SP, Brazil}
}

%
%


 

\begin{abstract}


This paper proposes \SConvTransform{}, a \Transform{} dialect extension that provides operations for optimizing 2D convolutions in MLIR. Its main operation, \SConvOp{}, lowers \Linalg{} convolutions into tiled and packed generic operations through a fully declarative transformation pipeline. The process is guided by a Convolution Slicing Analysis \del{(CSA)} that determines tile sizes and data layout strategies based on input and filter shapes, as well as target architecture parameters. \SConvOp{} handles edge cases by splitting irregular regions and adjusting affine maps where needed. All packing and tiling operations are derived from a parametric set of affine equations, enabling reusable and analyzable transformations. 

 Although functional correctness was the primary goal of this work, the experimental evaluation demonstrates the effectiveness of \SConvTransform{}, achieving good enough performance across different target architectures. Future work will focus on optimizing performance and porting to other target devices.
 
 When applied to standard convolution configurations, the generated code achieves up to 60\% of peak performance on Arm SME and 67\% on Intel AVX512. These results validate the benefit of combining static shape analysis with structured tiling and packing strategies within the MLIR \Transform{} dialect. Furthermore, the modular design of \SConvTransform{} facilitates integration with future extensions, enabling continued optimization of convolution workloads through MLIR's extensible compilation infrastructure.

\end{abstract}



\maketitle

\section{Introduction}

Convolution is a fundamental operation in image processing and \new{Deep Learning (DL)}, enabling the extraction of spatial features by applying filters across input tensors. In Convolutional Neural Networks (CNNs), convolutions typically dominate execution time due to their high arithmetic intensity and data reuse capabilities. Reducing convolution computational cost has motivated a wide range of optimization strategies, including data layout transformations, loop tiling, and architecture-specific scheduling.

Compiler frameworks such as MLIR~\cite{mlir} have recently emerged as powerful tools to express such optimizations at a high level, enabling systematic lowering of computations into efficient code for modern hardware. Within this context, optimizing convolutions presents particular challenges due to their multi-dimensional structure and the need for careful control over memory hierarchy and loop transformations.

This paper presents \textit{\SConvTransform{}}, a sliced convolution algorithm implemented as an extension of the MLIR \Transform{} dialect. It takes a \ConvOp{} as input and lowers it into one or more \GenericOp{} operations by applying a transformation pipeline that includes slicing, tiling, and packing. The transformation is guided by a static analysis pass, called Convolution Slicing Analysis (CSA), that determines how to partition the convolution, based on the target architecture capabilities (\eg{}, cache sizes), as well as how to organize memory layout to enhance data locality and overall performance.

Building upon previous work~\cite{sconv}, which used ad hoc rewrite patterns to express optimized convolution schedules, \SConvTransform{}\del{, or simply SConv,} introduces a more principled and composable approach by relying entirely on the Transform dialect infrastructure~\cite{transform}. It is organized as a sequence of transformation stages—edge case handling, slicing, tiling, packing, and affine map rewriting—implemented by the declarative operation \MlirSConv{}, enabling better integration with the MLIR ecosystem, as well as supporting reuse, modularity, and the evolution of transformation strategies.

Affine maps used in the packed operations are derived from a parametric set of equations formalized in Section~\ref{sec:packing-affine-modeling}. While convolutions that result in small tiles are currently handled without packing, future extensions will support automated padding to generalize the transformation pipeline. Altogether, \SConvTransform{} provides a flexible foundation for convolution optimization within MLIR, supporting both integration with diverse code generation backends and research on compiler transformations.

This paper is organized as follows. Sections~\ref{sec:sconv} to \ref{sec:packing-affine-modeling} describe the motivation, goals, and overall design of the \new{\SConvTransform{}}, including the edge-case handling and CSA-based scheduling strategy. Section~\ref{sec:implementation} details the implementation within the MLIR Transform dialect, highlighting the tiling, packing, and multipacking mechanisms for both \new{input stationary} and \new{weight stationary} schedules. Section~\ref{sec:experiments} presents the evaluation setup and performance results across representative convolution workloads. Section~\ref{sec:related} discusses related work in convolution optimization and MLIR-based transformations. Finally, Section~\ref{sec:conclusion} concludes the paper and outlines future directions.

\section{SConv Background} \label{sec:sconv}
This section introduces the core principles behind the  \textit{Sliced Convolution} (SConv) algorithm. It begins by reviewing the convolution and its notation, which is consistently used throughout the text. It then presents the SConv pipeline and details all of its steps. These elements form the conceptual basis for the implementation described in Sections~\ref{sec:porting-sconv} and \ref{sec:implementation}.

\subsection{Convolution Background} \label{sec:background}

A convolution is defined as the sliding of a set of filters (or kernels) over an input tensor to compute an output tensor. In the context of \new{Machine Learning (ML)}, these filters are also referred to as weights.

\begin{figure}[t]
    \centering
    \includegraphics[width=0.7\linewidth]{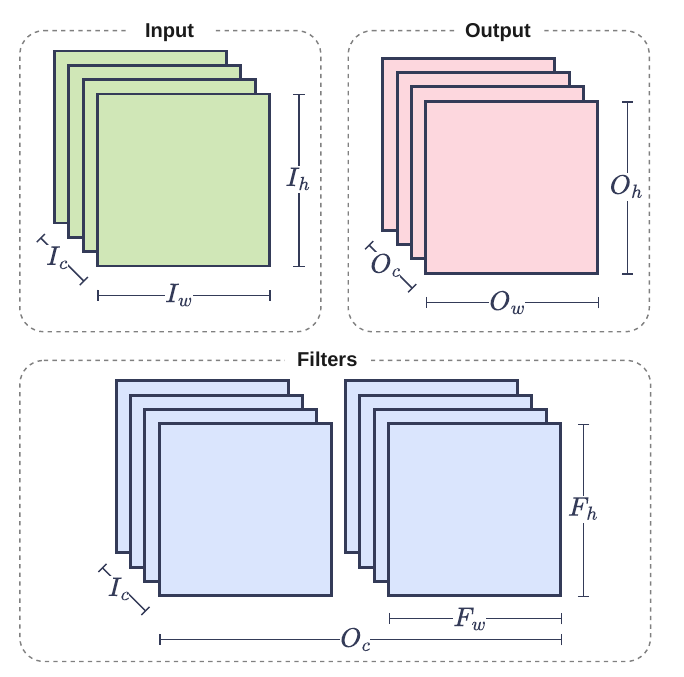}
    \caption{Convolution tensor layout}
    \label{fig:conv}
\end{figure}

The SConv algorithm supports 2D convolutions in the NCHW layout for input tensors and FCHW layout for filter tensors, as shown in \figurename~\ref{fig:conv}. Given an input tensor of shape $N \times I_c \times I_h \times I_w$ (batch size, input channels, height, width) and a filter tensor of shape $O_c \times I_c \times F_h \times F_w$ (output channels, input channels, filter height, filter width), the resulting output tensor has shape $N \times O_c \times O_h \times O_w$, where $O_h$ and $O_w$ depend on the convolution parameters: stride, padding, and dilation, respectively.

Each filter, of shape $I_c \times F_h \times F_w$, is applied across the spatial dimensions of the input. The projection of the filter onto a region of the input tensor is referred to as a \textit{window}. For each window position, the convolution computes a single scalar value through a weighted sum of the element-wise products between the filter and the corresponding input window.

These scalars are computed across all spatial positions (height and width), output channels, and batches, resulting in a nested loop structure over the relevant dimensions. The above discussed paramenters $N$, $I_c$, $O_c$, $I_h$, $I_w$, $F_h$, $F_w$, $O_h$, and $O_w$ are consistently used in Section~\ref{sec:packing-affine-modeling} and throughout Section~\ref{sec:implementation} to define affine maps and transformations in MLIR.

\subsection{The SConv Algorithm} \label{sec:sconv-algorithm}

SConv\del{ (Sliced Convolution) }~\cite{sconv} is a direct convolution algorithm designed for ML compilers and portability across architectures, including vector~\cite{intel} and matrix~\cite{cal_2024_kim,mma,helloSME} extensions. It leverages architectural information and vendor-specific portable microkernels to boost computational performance tailored to the available hardware. Additionally, it optimizes memory and cache usage by employing a tiling strategy that efficiently slices and schedules the convolution input, filter/weight, and output tensors. SConv is only suitable for convolutions in the NCHW format.

\begin{figure}
    \centering
    \includegraphics[width=0.8\linewidth]{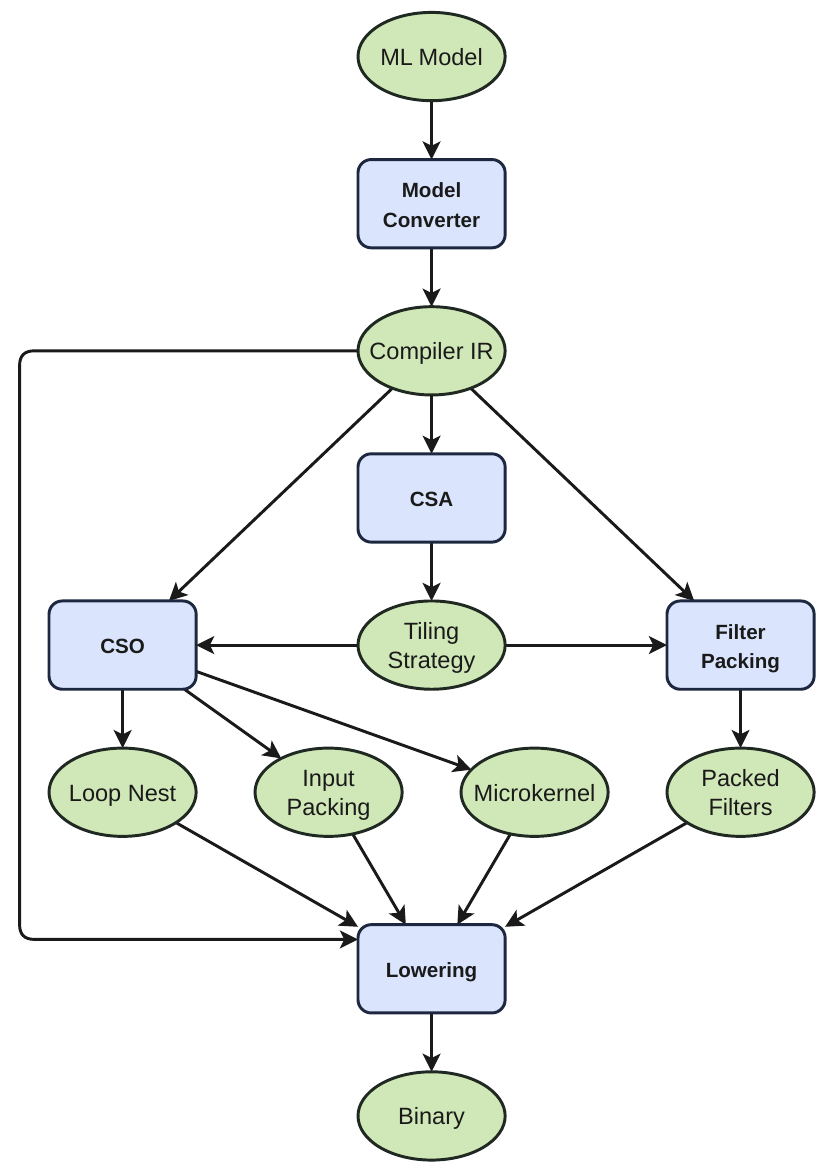}
    \caption{Simplified compilation flow for a machine learning compiler using SConv to optimize convolutions.}
    \label{fig:generic-flow}
\end{figure}

The algorithm is structured as a sequence of compile-time optimization passes targeting convolution code generation. \figurename~\ref{fig:generic-flow} illustrates the SConv pipeline, with optimization passes depicted as blue blocks and their intermediate outputs as green ellipses, within a typical ML compilation workflow. While applicable to any convolution workload, SConv is especially optimized for \new{Artificial Intelligence (AI)} model compilation. 

The process begins with the compiler \new{Intermediate Representation (IR)} generated from the model, where convolution operations are explicitly represented. At this stage, sufficient information is available to initiate the SConv pipeline, which consists of the Convolution Slicing Analysis (CSA) pass, responsible for cache blocking analysis, and the Convolution Slicing Optimization (CSO) pass, responsible for generating efficient macrokernel code.

\begin{figure}
    \centering
    \includegraphics[width=0.9\linewidth]{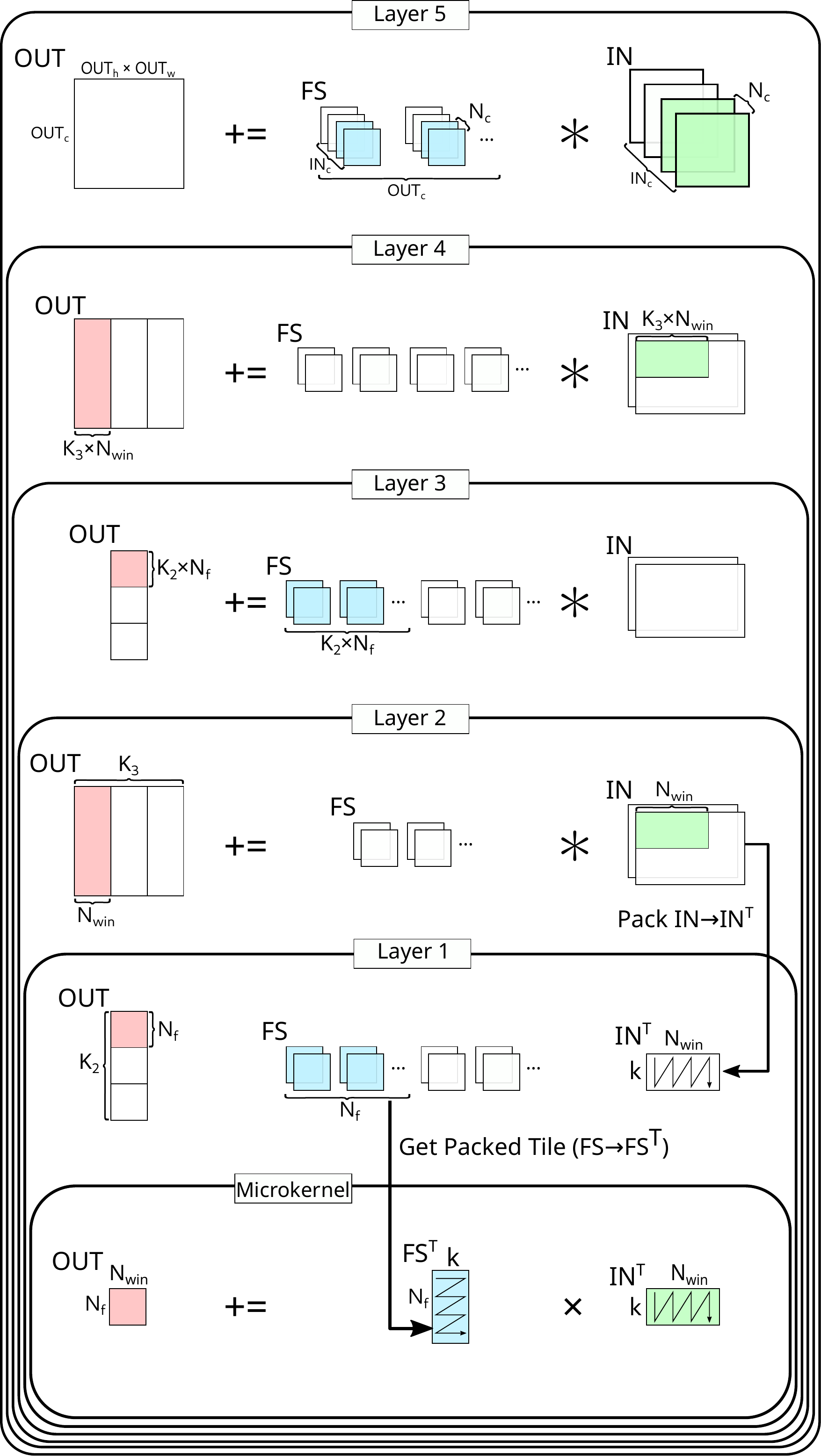}
    \caption{Convolution macro-kernel as generated by the Convolution Slicing Optimization pass of the SConv algorithm.}
    \label{fig:cso}
\end{figure}

The CSO, illustrated in \figurename~\ref{fig:cso}, is the code generation phase that constructs the macrokernel loop nest. It coordinates tiling, packing, and scheduling to efficiently feed the microkernel, guided by parameters produced by the preceding CSA pass. The CSO macrokernel consists of five layers surrounding the microkernel, each layer tiling in one dimension. Layer 5 tiles both the input and filter set tensors in the channel dimension $I_c$, with a CSA-defined parameter $N_c$ based on the L1 cache size. From this point, there are two levels of tiling in the same dimensions: number of windows ($N_{win}$) and number of filters ($N_f$). Both of these are dependent on architecture and the microkernel, as described later in this subsection. The first level of tiling, done in Layers 4 and 3, splits the tensors into elements that correspond to sets of tiles of size $K_2$ and $K_3$, defined by CSA based on the sizes of the L2 and L3 caches, while the second level of tiling, done by layers 2 and 1, splits a single tile out of each set. The scheduling strategy calculated by CSA determines the order between the filter and the input tiling in both levels.

CSA defines the convolution \textit{Tiling Strategy}, which includes tile size selection, hierarchical tile distribution, and tile scheduling. These decisions are guided by cache hierarchy constraints (cache size, cache line) and microkernel resource requirements. Tile sizes are selected to ensure that a single tile from the input, filter, and output tensors can simultaneously fit within the L1 cache. CSA further partitions these tiles across deeper cache levels and schedules their evaluation to promote maximum data reuse. The scheduling strategy, chosen via a cost model that estimates memory traffic, can be either \textit{Input Stationary} (IS) or \textit{Weight Stationary} (WS), depending on which tensor benefits most from reuse. 

Tiling space exploration for cache blocking is performed by CSA, which heuristically determines appropriate values for $N_c$, $K_2$, and $K_3$ to maximize data reuse and efficiently utilize the memory hierarchy. The parameter $N_c$ denotes the number of input channels processed per tile, while $K_2$ and $K_3$ specify the number of tiles of a given type (input or filter) that are retained in the L2 and L3 caches, respectively. The selected scheduling strategy determines which tile type is prioritized at each cache level, improving data locality and minimizing redundant memory traffic. In short, to reduce data re-use, it is desirable, at each microkernel call, to have slices of the larger tensor (input or filter) stationary at L1 cache and design the scheduler to touch it once. For additional details of the inner workings of CSA, the reader should refer to \cite{sconv}.

To maximize hardware throughput, SConv~utilizes an architecture-specific outer-product microkernels to compute individual output tiles. These microkernels leverage specialized \new{Instruction Set Architecture (ISA)} extensions, such as IBM MMA~\cite{mma} and Intel AVX512. The outer-product formulation achieves high computational density by producing $n^2$ outputs from $2n$ inputs, leading to its broad adoption in high-performance linear algebra libraries and its incorporation in modern ISAs such as IBM POWER10 MMA. Each microkernel call performs a sequence of outer-products between input tensor windows and filters, thus requiring that the input be decomposed into \textbf{windows} before its invocation. 

Hardware constraints dictate the number of windows and filters that can be computed per microkernel invocation. This granularity is characterized by the \textit{microkernel shape}, defined as $N_{win} \times N_f$, where $N_{win}$ represents the number of output windows (\ie{}, spatial positions) computed per call, and $N_f$ denotes the number of filters processed at once. These parameters are used to define the tile layout and the scheduling strategy during code generation. The microkernel also uses it to specify its internal register allocation strategy.

With a convolution-specific tiling strategy in place, SConv proceeds to reorganize tensors for calling the microkernel, a procedure named \textit{Packing}, which serves as the link between tile-level memory layout and microkernel execution. To achieve top performance, microkernels are designed to explore memory locality as much as possible. Packing reorganizes data for efficient access based on the microkernel’s specification, which is inherently architecture-dependent. As such, the microkernel expects its input/filter parameters to be sequentially stored (ie, \textit{packed}) in memory. In SConv, packing happens on-the-fly just before tile computation. This method is called \textit{Packing-on-Demand} and reduces the overhead typically seen in Im2Col + BLAS approaches, which suffer from redundant packing and coarse-grained tiling.

The CSA tiling strategy parameters \(N_c\), \(K_2\), and \(K_3\), along with the microkernel parameters \(N_{win}\) and \(N_f\), are referenced consistently throughout the implementation discussion in Sections~\ref{sec:porting-sconv} and~\ref{sec:implementation}.

\section{Porting SConv to MLIR} \label{sec:porting-sconv}

SConv's parametrized approach to architecture information and compiler-centric design favors its integration within the MLIR toolchain, specifically for \new{AI} workloads. In this approach, during a typical compiler retargetting build, the programmer provides microkernel and architecture information to SConv, thus enabling the lowering process to be predominantly platform-agnostic. The specific mechanisms for providing this information are discussed in Section~\ref{sec:api}.

When integrating SConv into MLIR, the compilation flow requires some significant changes when compared with a traditional ML flow.
Contrary to a traditional \new{ML} compiler pipeline, where convolution is eventually reduced to library generation with hard-coded implementations, SConv is a \texttt{Structured Transform} pass within the MLIR lowering framework that applies successive transformations to the original convolution operator, leading to the progressive generation of a macrokernel and microkernel. This integration enables SConv to be utilized in both AI and non-AI workflows, including image processing and engineering applications, provided that the appropriate dialects and operations are employed. This integration builds upon an RFC originally proposed to the LLVM community in 2023\footnote{RFC post: \url{https://discourse.llvm.org/t/rfc-optimized-convolution-for-mlir/69454}}.

\subsection{MLIR Transform Dialect}

The \Linalg{} dialect serves as the entry point for \SConvTransform{} passes, as it is typically the final dialect in which convolution remains a high-level, semantically transparent operation. This lowering phase performs code generation for linear algebra algorithms, including convolution. Consequently, it represents the final stage where optimization opportunities can be readily identified and the earliest stage in which such transformations can be effectively applied.

To enable the generation of complex algorithm implementations at compile time and avoid hardcoded, target-specific library code, the MLIR infrastructure introduced the \Transform{} dialect~\cite{transform}. This dialect enables the construction of reusable templates that apply both high-level transformations, such as tiling, as well as lower-level optimizations, such as canonicalization, to the payload IR, when applicable. It is frequently employed to implement algorithmic lowering for named \Linalg{} operations, including widely-used techniques like the Im2Col + GEMM method~\cite{im2col} and Winograd convolution~\cite{winograd}. As such, the \Transform{} dialect was a natural integration point for the SConv pipeline.

Unlike previous implementations~\cite{sconv}, which relied on imperative logic code generation or external library calls tailored to specific backends, the integration of SConv into the \Transform{} dialect demanded a complete reexpression of the algorithm using declarative MLIR constructs. This rearchitecture required careful decomposition of each transformation step (slicing, tiling, and packing) into composable operations compatible with the \Linalg{}, \Affine{}, \SCF{}, and \Tensor{} dialects. The result is a modular and analyzable pipeline, fully embedded within MLIR’s transformation infrastructure.

\subsection{Transform Operation Structure}
The CSO stage of the SConv pipeline (detailed in Section \ref{sec:sconv-algorithm}) is implemented as a sequence of transformation passes applied to each convolution operation of the model that processes its NCHW input. These passes are specific to the implementation and are described in detail in Section~\ref{sec:implementation}. After each transformation, the generated code remains valid and can be lowered through the standard LLVM compilation flow, preserving correctness throughout.

By decomposing the CSO code generation algorithm into discrete, modular passes, this approach improves generality and portability. The architectural parameters required for each transformation are explicitly exposed and user-defined, decoupling the routine from any specific compilation target. Instead, it operates entirely on configuration provided at the API level, as detailed in Section~\ref{sec:api}.

In this work, a microkernel is also provided by the programmer, as the current version does not support automatic microkernel generation by the compiler, which will be addressed in future work. Further details on microkernel lowering are presented in Section~\ref{sec:lowering}.

\subsection{SConv Transform Compatibility}\label{sec:compat}

The implementation of SConv as a \Transform{} dialect extension is called \SConvTransform{}. Its core logic is encapsulated in the \SConvOp{} C++ class, which serves as the main operation of \SConvTransform{}. In the MLIR Transform dialect IR, \SConvOp{} operation is invoked through \MlirSConv{}.

In addition to the high-level structural modifications needed for design \SConvTransform{}, specific implementation steps had to be adapted to better align with the MLIR infrastructure and to enable further optimizations during lowering. Moreover, many required layout transformations have been detected to be incompatible with the standard structure of a \Linalg{}  convolution operator. These constraints and their corresponding solutions are detailed later in this paper. Overall, to address them, \SConvTransform{} implementation extensively utilizes \MlirGeneric{} operations. While this reliance constrains certain aspects of code generation, it aligns better with the philosophy of the \Transform{} dialect by producing high-level, concise, and generic code. Such code affords greater flexibility for subsequent transformations, optimizations, and lowering as a cohesive unit once the whole transformation sequence is complete.

The initial step in generating the loop nest that implements SConv (Figure \ref{fig:cso}), for the appropriate tiling strategy, involves collapsing the spatial dimensions \(H\) and \(W\) of both the input and the output tensors into a single dimension. Upon completion of the transformation, these shapes are subsequently expanded to match the original NCHW specifications. This linearization is necessary due to limitations in the MLIR tiling infrastructure when handling \Linalg{} dialect operations. Specifically, SConv treats the output spatial dimensions as a set of windows computed across rows; consequently, tiling must be performed along a single dimension corresponding to these windows rather than independently across rows and columns. To correctly handle edge cases, the spatial dimensions of the input tensor are also collapsed.

Throughout the entire \SConvTransform{} flow, the \MlirGeneric{} convolution operation is progressively transformed into a set of nesting loops that execute a convolution microkernel, while maintaining its generic representation. The packing steps are likewise implemented using \MlirGeneric{} operations rather than explicit loop nests, necessitating a formal modeling of the packing procedure for both input and filter tensors via affine indexing maps. This modeling is detailed in Section~\ref{sec:packing-affine-modeling}.

The tiling process can generate edge cases arising from the microkernel parameters \(N_{win}\) and \(N_f\) in layers 1 and 2 of \figurename~\ref{fig:cso}, as well as from the CSA parameters \(N_c\), \(K_2\), and \(K_3\) determined by the CSA tiling strategy. To maintain scalability and preserve the use of the \MlirGeneric{} structure during code generation, these edge cases are addressed via split transformations that \textit{peel} the epilogue of a given dimension from the steady-state iteration space. Although this approach results in an exponential increase in the number of code paths to cover all combinations, each path follows an identical lowering process, preventing excessive complexity. An additional optimization, not implemented in this release, would integrate the split transformations within the tiling process to avoid redundant loop generation.

Implementation details of \SConvTransform{}, including examples of generated code, are provided in Section~\ref{sec:implementation}.

\section{SConv Packing}
\label{sec:packing-affine-modeling}

As shown in Figure~\ref{fig:generic-flow}, the SConv packing step is divided into \textit{Filter Packing} and \textit{Input Packing}, the order of which is determined by the scheduling defined by the CSA tiling strategy. Both have very similar final layouts, but differ in their starting layouts and replication needs, so they are modeled separately. The tiling process precedes packing, so the tiles are already separated with the correct dimensions, as detailed in the respective subsections below.

This section describes how packing is modeled using affine maps within \MlirGeneric{} operations. The implementation also incorporates an optimization termed \textit{Multipacking}, wherein multiple tiles are packed simultaneously at a higher level. This optimization is applied to the innermost packing performed in Layer 1 of \figurename~\ref{fig:cso}, which is hoisted to Layer 3 as the packing of \(K_2\) tiles designed to fill the L2 cache.

The SConv microkernel has a shape of \(N_f \times N_{win}\), meaning each packed tile type (filter and input tensors) spans one of these output dimensions and includes one or more reduction dimensions that comprise the remainder of the tile.

\subsection{Filter Packing}
Filter packing is the simpler of the two packing procedures used in SConv, as it involves a straightforward reordering of filter elements without replication. In this implementation, elements are stored in order but accessed out of order. Filter packing can be performed statically when filter data is embedded in the generated code, although this applies only to specific workloads. Equation~\ref{eq:filter_tile_shape} gives the shape of the filter set tile, while the shape of the packed tile is shown in Equation~\ref{eq:filter_packed_tile_shape}.

\begin{align}
    \text{Filter tile shape:} \quad & N_f \times N_c \times F_h \times F_w \label{eq:filter_tile_shape} \\
    \text{Packed tile shape:} \quad & N_c \times F_h \times F_w \times N_f \label{eq:filter_packed_tile_shape}
\end{align}

In Filter Packing, the packing corresponds to a dimension reorder analogous to a transpose operation. Although this could be modeled using an \MlirTranspose{} operation, the \MlirGeneric{} approach demonstrated superior performance. In this model, the operation iterates over the dimensions of the packed tile, loading elements in sequence and storing them directly into the output.

When applying the Multipacking optimization, an additional dimension \(N_t\) (corresponding to either \(K_2\) or \(K_3\)) is introduced into the packed structure to represent the aggregation of packed tiles. This introduces a new induction variable \(iN_t\), which, together with the induction variable \(iN_f\) spanning the filter dimension, is used to compute the starting index of the tile within the original filter set, denoted as \(iT_f\) (see Equation~\ref{eq:filter_multipacking}). The variable \(iT_f\) replaces \(iN_f\) for indexing the filter dimension. The filter set tile is also much larger, containing all the filters for the \(N_t\) tiles.

\begin{align}
    iT_f = iN_t \cdot N_f + iN_f \label{eq:filter_multipacking}
\end{align}

\subsection{Input Packing}
Input packing is more complex due to the necessity of replication, resulting in more intricate equations and affine maps to model the packing process accurately. Linearization of input tensor dimensions and handling of edge cases further increase complexity and impact performance. The shape of the input tensor tile is given in Equation~\ref{eq:input_tile_shape}, while the shape of the packed tile is shown in Equation~\ref{eq:input_packed_tile_shape}. For simplicity, these equations assume a batch size of one, as the batch dimension does not affect SConv.

\begin{align}
    \text{Input tile shape:} \quad & N_c \times F_h \times (N_{win} + F_w - 1) \label{eq:input_tile_shape} \\
    \text{Packed tile shape:} \quad & N_c \times F_h \times F_w \times N_{win} \label{eq:input_packed_tile_shape}
\end{align}

The elements are stored in order and are predominantly loaded sequentially, processing \(N_{win}\) elements at a time while skipping certain elements at the beginning or end of each row, due to how the filter is projected onto the input tensor to form windows. $\lfloor \frac{F_w}{2}\rfloor$ elements are skipped at each end of the row.

Data replication occurs between windows due to overlap when the \textit{stride} is small. When projecting the filter onto the input tensor, if the stride is smaller than $\lfloor \frac{F_w}{2}\rfloor$ in the width dimension or $\lfloor \frac{F_h}{2}\rfloor$ in the height dimension, some elements will be the same between two windows. The reader should notice that \textit{Vector-based Packing}, an optimized vector register-based packing algorithm proposed in \cite{sconv}, is not used in this initial release of \SConvTransform{}.

The input packing operation iterates over the dimensions of the packed tile, using the induction variables \(iN_c, iF_h, iF_w, iN_{win}\). The input tensor is accessed with indices \(iN_c\) and \(iT_{hw}\), where \(T_{hw}\) denotes the linearized spatial dimension of a single tile after splitting from the complete convolution input tensor. The original spatial dimensions of the tile are \(T_{h}\) and \(T_{w}\) (height and width). Variable \(iN_c\) can be used directly in all cases described below to access the input channel dimension, as this dimension has the same size and layout between the input tensor slice and the packed tile.

The simplest variation of input packing occurs when all data to be packed lies within the same input tensor row. In this case, \(iF_h\) is the sole variable used for row calculation, while the column calculation depends exclusively on \(iF_w\) and \(iN_{win}\), as detailed in Equations~\ref{eq:input_ihtile}, \ref{eq:input_iwtile}, and \ref{eq:input_ihwtile}. The filter row \(iF_h\) determines which tile row to access, whereas the current window \(iN_{win}\) and filter column \(iF_w\) determine the tile column, with dilation and stride parameters applied when needed.

\begin{align}
    iT_h &= iF_h \cdot dilationH \label{eq:input_ihtile} \\
    iT_{w} &= iN_{win} \cdot strideW + iF_w \cdot dilationH \label{eq:input_iwtile} \\
    iT_{hw} &= iT_h \cdot T_w + iT_{w} \label{eq:input_ihwtile}
\end{align}

However, in the general case where row breaks may occur within the tile, input packing cannot be performed independently of the original tensor, even if the tile has been previously split. This is because the handling of the end of the input tensor rows differs from the rest. Consequently, information about the current column within the complete tensor is required to determine where rows should be skipped. Such information can be derived from the iterators of the outer and inner tiling loops over the output spatial dimension.

Consider \(iO_{out}\) and \(iO_{in}\) as the outer and inner loop iterators, respectively. Their combination defines \(iT_{s}\), which corresponds to the starting position of the tile (Equation~\ref{eq:input_tile_start}). The current overall window of the tensor, \(iO_{hw}\) (Equation~\ref{eq:input_window}), is computed by combining \(iT_{s}\) with the current window index within the tile. This value is used to determine the current row and column of the complete tensor that needs to be accessed. Specifically, the row is given by the integer division of the window index by the number of windows per row (\ie, the output width \(O_w\)), while the column corresponds to the remainder of this division. Using \(iT_{s}\), the row and column indices for tile access can be derived. This process is detailed in Equations~\ref{eq:input_htile_full} and \ref{eq:input_wtile_full}, while \(iT_{hw}\) is still calculated as in Equation~\ref{eq:input_ihwtile}.

\begin{align}
  iT_{s} &= iO_{out} + iO_{in} \label{eq:input_tile_start} \\
  iO_{hw} &= iT_{s} + iN_{win} \label{eq:input_window} \\
  iT_h &= \left(\frac{iO_{hw}}{O_w} - \frac{iT_{s}}{O_w}\right) \cdot strideH + iF_h \cdot dilationH \label{eq:input_htile_full} \\
  iT_{w} &= \left(iO_{hw} \bmod O_w - iT_{s} \bmod O_w\right) \cdot strideW + iF_w \cdot dilationW \label{eq:input_wtile_full}
\end{align}

\(iT_{s}\) can be computed outside the \MlirGeneric{} operation, given that it is invariant with respect to the induction variables within it.

\subsubsection{Multipacking}
For the Multipacking optimization, a new dimension \(N_t\) (corresponding to either \(K_2\) or \(K_3\)) is introduced into the packed structure, representing the dimension that covers the packed tiles. Each iteration over this dimension skips a full tile, which modifies the calculation of the current window to be accessed, as shown in Equation~\ref{eq:input_window_multipacking}. Additionally, since multipacking operates on a larger tile before the inner tiling loop, only the outer loop iterator is used in \(iT_{s}\), as expressed in Equation~\ref{eq:input_tile_start_multipacking}.

\begin{align}
    iT_{s} &= iO_{out} \label{eq:input_tile_start_multipacking} \\
    iO_{hw} &= iT_{s} + iN_t \cdot N_{win} + iN_{win} \label{eq:input_window_multipacking}
\end{align}

The remaining indices are computed as before, using Equations~\ref{eq:input_htile_full}, \ref{eq:input_wtile_full}, and \ref{eq:input_ihwtile}.

\subsubsection{Edge Packing}
The final variation of input packing addresses edge cases in the linearized output spatial dimension, where the operation is divided into a steady-state and an epilogue phase, each with distinct tiling loops and iterators. Because the split can occur mid-row and the epilogue may span multiple rows, additional information is required for \(iT_{s}\). Specifically, in the epilogue phase, the absolute window also depends on the offset from the tensor start to the operation start, denoted as \(E_{off}\).

\begin{align}
    iT_{s} = iO_{out} + iO_{in} + E_{off} \label{eq:input_tile_start_edge}
\end{align}

The following section demonstrates how these decisions are implemented within MLIR, progressing from high-level convolution to a fully transformed and scheduled kernel.

\section{SConvTransform Implementation}
\label{sec:implementation}

\begin{figure*}[ht]
    \centering
    \includegraphics[width=1.0\linewidth]{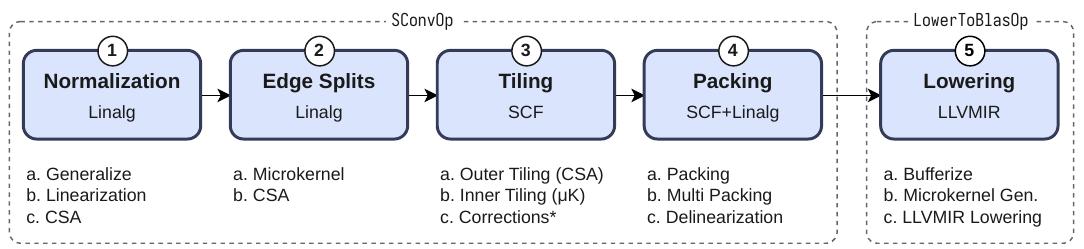}
    \caption{SConvTransform Compilation Flow}
    \label{fig:flow}
\end{figure*}

The overall structure of the transformation pipeline implemented by \SConvTransform{} is illustrated in Figure~\ref{fig:flow}, where \SConvOp{} orchestrates the execution of all pipeline stages. Each stage corresponds to a set of tasks identified by the letters under the boxes in the figure.

The flow begins executing task~\circled{1a} at stage 1, where specific convolution operations (\eg{}, \ConvOp{}) are identified in the input IR and transformed into a standardized \GenericOp{} form suitable for subsequent transformations. This task is detailed in Section~\ref{sec:overview}, including the application of the shape collapsing operation---represented in the figure as Linearization---which flattens the spatial dimensions of input and output tensors~\circled{1b}. The process then progresses to~\circled{1c}, where the scheduling and tiling parameters are calculated using CSA analysis. Section~\ref{sec:schedule} describes how architectural constraints and convolution properties drive this decision.

Before applying tiling, the transformation analyzes the potential edge cases in \circled{2}, as explained in Section~\ref{sec:edge-cases}. This step determines whether the input or filter domains must be split to accommodate irregular sizes~\circled{2a} or scheduling constraints~\circled{2b}.

At~\circled{3}, loop transformations and two-level tiling are applied to the regularized \GenericOp{} operations representing convolutions (hereafter referred to simply as kernels), producing structured loop nests compatible with microkernel execution. The implementation of this transformation, along with its structure and IR output, is covered in Section~\ref{sec:tiling}.

In~\circled{4}, the transformation reorganizes the data layouts through packing and multipacking. These operations ensure that inputs and filters are reshaped and aligned for efficient memory access and vectorized computation. The details are presented in Section~\ref{sec:packing}.

Finally, in~\circled{5} the transformation will lower operations on the Tensor dialect to operations on the MemRef dialect, a step called bufferization. Then, it swaps the existing generic microkernel for a function call to the BLAS library microkernel. As mentioned before, in its current release, \SConvTransform{} does not generate code for the microkernel. 

The following subsections describe each of the stages of Figure~\ref{fig:flow} in detail, illustrating how high-level convolution operations are progressively lowered into tiled and packed kernels optimized for performance.

\subsection{Convolution Normalization (stage 1)}
\label{sec:overview}

The first stage of the \SConvOp{} \circled{1} is shown in Figure~\ref{fig:flow} as tasks~\circled{1a} and~\circled{1b}. In~\circled{1a}, named convolution operations such as \MlirConv{} are detected in the input IR. A typical example is shown in Listing~\ref{lst:named-conv}, where the convolution is defined on input, weight, and output tensors using the NCHW and FCHW layouts, with explicit stride and dilation attributes.

\begin{figure}[t]
\begin{lstlisting}[language=mlir, caption={Named convolution.}, label={lst:named-conv}, numbers=left]
module {
  func.func @conv_2d_nchw_fchw(%in: tensor<1x32x152x152xf32>,
                               %wei: tensor<256x32x3x3xf32>,
                               %out: tensor<1x256x75x75xf32>) -> tensor<1x256x75x75xf32> {
    %res = linalg.conv_2d_nchw_fchw
      {dilations = dense<[1,1]>, strides = dense<[2,2]>}
      ins(%in, %wei : tensor<1x32x152x152xf32>, tensor<256x32x3x3xf32>)
      outs(%out : tensor<1x256x75x75xf32>)
      -> tensor<1x256x75x75xf32>
    return %res : tensor<1x256x75x75xf32>
  }
}
\end{lstlisting}
\end{figure}

In~\circled{1b}, these named convolutions are transformed into a standardized \GenericOp{} form. This transformation includes the shape collapsing operation—shown in the figure as Linearization—which flattens the spatial dimensions of both input and output tensors, enabling fine-grained affine analysis. Listing~\ref{lst:generic-conv} shows the resulting kernel, where the convolution is expressed as an \MlirGeneric{} operation with custom indexing maps and iterator types.

This generic representation exposes the convolution’s complete loop structure—parallel and reduction loops, affine indexing logic, and tensor shapes, making it suitable for \new{CSA} and later stages of tiling and packing.

\subsubsection{Schedule Analysis and Tiling Strategy (CSA)}
\label{sec:schedule}

Once the convolution has been generalized (as shown in Listing~\ref{lst:generic-conv}), a \new{CSA} is performed \circled{1c} to determine the appropriate tiling schedule. The analysis selects Input Stationary (IS) or Weight Stationary (WS) scheduling, and computes tile sizes along three critical dimensions: the input channel tile size, $N_c$, and two tile sizes in non-reduction dimensions:  $K_2$, for the output channel dimension ($O_c$), and $K_3$, for the flattened spatial dimension of the output tensor ($O_h \times O_w$). These tile sizes are chosen to maximize data reuse and locality, aligning with the L2 and L3 cache sizes. To determine these values, the CSA analysis combines architectural configuration, microkernel characteristics, and convolution dimensions in a cost-driven model that guides schedule selection and tiling granularity.

\begin{figure}[t]
\begin{lstlisting}[language=mlir, caption={Generic convolution.}, label={lst:generic-conv}, numbers=left]
%0 = linalg.generic
  {indexing_maps = [
      affine_map<(d0, d1, d2, d3, d4, d5) -> 
        (d0, d3, ((d2 floordiv 75) * 2 + d4) * 152 + (d2 mod 75) * 2 + d5)>,
      affine_map<(d0, d1, d2, d3, d4, d5) -> (d1, d3, d4, d5)>,
      affine_map<(d0, d1, d2, d3, d4, d5) -> (d0, d1, d2)>],
   iterator_types = ["parallel", "parallel", "parallel", "reduction", "reduction", "reduction"]}
  ins(%collapsed, %arg1 : tensor<1x32x23104xf32>, tensor<256x32x3x3xf32>)
  outs(%collapsed_0 : tensor<1x256x5625xf32>) {
  ^bb0(%in: f32, %in_1: f32, %out: f32):
    %1 = arith.mulf %in, %in_1 : f32
    %2 = arith.addf %1, %out : f32
    linalg.yield %2 : f32
} -> tensor<1x256x5625xf32>
\end{lstlisting}
\end{figure}

Along with each tile size, CSA also computes the corresponding remainder values $R_{N_c}$, $R_{K_2}$, and $R_{K_3}$. Each remainder represents an edge case where the respective dimension is not evenly divisible by the selected tile size:
$$
R_{X} = \text{dimension size} \bmod \text{tile size for that dimension}.
$$

A remainder value of zero indicates that the dimension fits an integer number of tiles; a nonzero value signals that an extra partial tile will be needed.
For the example presented in Listing~\ref{lst:generic-conv}, CSA selected the input stationary schedule, with the following parameters:

\smallskip
\begin{center}
$K_2=32$,\quad $R_{K_2}=0$,\quad $K_3=87$,\quad $R_{K_3}=3$,\quad $N_c=32$,\quad $R_{N_c}=0$
\end{center}
\smallskip

This configuration means that tiling along $K_2$ (filter output channels) and $N_c$ (input channels) produces no remainders. However, tiling along $K_3$ does not divide evenly into the output spatial dimension $O_h \times O_w = 75 \times 75 = 5625$, resulting in a remainder of $R_{K_3} = 3$. This constitutes an edge case for $K_3$, which must be handled explicitly.

Section~\ref{sec:edge-cases} illustrates how the transformation pipeline addresses such edge cases by applying recursive splitting. It shows both the edge case in \(K_3\) and a structural edge case on the input domain (when \(O_h\)~$\times$~\(O_w\) is not divisible by \(N_{win}\)). In each case, it shows how the original kernel is split into regular and remainder regions, and how each resulting kernel is transformed independently.

\subsection{Edge Case Handling (stage 2)}
\label{sec:edge-cases}

Following CSA analysis in \circled{1c}, the implementation proceeds to handling edge cases\circled{2}. It first checks for edge cases in the input and filter domains~\circled{2a}. These structural edge cases occur when the total number of output spatial positions (\(O_h\) $\times$ \(O_w\)) is not divisible by the number of windows (\(N_{win}\)), or when the number of filters (\(N_c\)) is not divisible by the number of filters per tile (\(O_c\)). Such misalignments result in a convolution that cannot be evenly tiled. In these cases, the operation is split into a main kernel and a remainder kernel. The latter, being smaller than the tiling unit, is currently left untiled and unpacked, with only a canonicalization of its indexing. Support for padding and full integration of these kernels into the optimization pipeline is planned for future releases.

\sconvpar{Edge case on the input domain.} 
In the example of Listing~\ref{lst:generic-conv}, the spatial output domain $O_h \times O_w$ has a dimension of $75 \times 75 = 5625$, which is not divisible by the number of windows defined for tiling ($N_{win}=16$). This leads to an initial edge case along the flattened spatial dimension. To handle it, the operation is split at index $5616$, isolating a small tail of $9$ positions.

The result is two independent \MlirGeneric{} kernels. The main kernel, shown in Listing~\ref{lst:main-kernel}, processes $5616$ spatial positions and proceeds normally through tiling and packing. The remaining kernel, with only $9$ positions, as shown in Listing~\ref{lst:remainder-kernel}, is smaller than $N_{win}$ and thus bypasses the complete transformation pipeline. For this kernel, only a minimal correction on affine maps is applied. Future versions of \SConvOp{} may implement padding to align such kernels with the window size and use the same optimization steps.

\begin{figure}[t]
\begin{lstlisting}[language=mlir, caption={Main kernel for input domain edge case.}, label={lst:main-kernel}, numbers=left]
%0 = linalg.generic {
  indexing_maps = [
    affine_map<(d0, d1, d2, d3, d4, d5) ->
      (d0, d3, ((d2 floordiv 75) * 2 + d4) * 152 + (d2 mod 75) * 2 + d5)>,
    affine_map<(d0, d1, d2, d3, d4, d5) ->
      (d1, d3, d4, d5)>,
    affine_map<(d0, d1, d2, d3, d4, d5) ->
      (d0, d1, d2)>],
  iterator_types = ["parallel", "parallel", "parallel", "reduction", "reduction", "reduction"]
}
ins(%extracted_slice, %extracted_slice_1 : tensor<1x32x22933xf32>, tensor<256x32x3x3xf32>)
outs(%extracted_slice_2 : tensor<1x256x5616xf32>) {
^bb0(%in: f32, %in_7: f32, %out: f32):
  %2 = arith.mulf %in, %in_7 : f32
  %3 = arith.addf %2, %out : f32
  linalg.yield %3 : f32
} -> tensor<1x256x5616xf32>
\end{lstlisting}
\end{figure}

\begin{figure}[t]
\begin{lstlisting}[language=mlir, caption={Remaining kernel for input domain edge case.}, label={lst:remainder-kernel}, numbers=left]
%1 = linalg.generic {
  indexing_maps = [
    affine_map<(d0, d1, d2, d3, d4, d5) -> 
      (d0, d3, ((d2 floordiv 75) * 2 + d4) * 152 + (d2 mod 75) * 2 + d5)>,
    affine_map<(d0, d1, d2, d3, d4, d5) -> 
      (d1, d3, d4, d5)>,
    affine_map<(d0, d1, d2, d3, d4, d5) -> 
      (d0, d1, d2)>],
  iterator_types = ["parallel", "parallel", "parallel", "reduction", "reduction", "reduction"]
}
ins(%extracted_slice_3, %extracted_slice_4 : tensor<1x32x323xf32>, tensor<256x32x3x3xf32>)
outs(%extracted_slice_5 : tensor<1x256x9xf32>) {
  ^bb0(%in: f32, %in_7: f32, %out: f32):
    %2 = arith.mulf %in, %in_7 : f32
    %3 = arith.addf %2, %out : f32
    linalg.yield %3 : f32
} -> tensor<1x256x9xf32>

\end{lstlisting}
\end{figure}

In contrast, edge cases arising during the application of the CSA tiling strategy~\circled{2b}—specifically for \(N_c\), \(K_3\), and \(K_2\)—are treated differently. These cases arise when the input channel dimension or the reduction window (\(I_c\) $\times$ \(F_h\) $\times$ \(F_w\)) is not divisible by \(N_c\), or when \(O_h\) $\times$ \(O_w\) and \(O_c\) are not divisible by \(K_2\) or \(K_3\), respectively. A recursive and ordered splitting process is applied, starting from \(K_2\), then \(K_3\), and finally \(N_c\). Each split produces two kernels: one aligned with the tile size and one containing the remainder. This process may be nested; for instance, if both \(K_2\) and \(K_3\) produce edge cases, three kernels result, each of which is fully transformed with tiling and packing. This recursive decomposition ensures coverage of the full iteration space with localized transformations tailored to the fit of the tiles.

\sconvpar{Edge case on \(K_3\).}
Returning to the example on Listing~\ref{lst:main-kernel}, the 5616-element main kernel is then inspected for alignment with the \(K_3\) tile size selected by CSA. In this case, \(K_3\) was set to 87, which does not divide 5616 evenly: $5616 = 87 \times 64 + 48$.

The 5616-element main kernel is further split into: (1) A new main kernel, as shown in Listing~\ref{lst:tiled-kernel}, covering 5568 elements (divisible by 87), which proceeds through multi-level tiling and packing; (2) A remaining kernel with 48 elements, as shown in Listing~\ref{lst:extra-kernel}, which, being larger than \(N_{win}\), also follows the full tiling and packing pipeline.

This recursive splitting strategy ensures that all portions of the convolution domain are either optimized or safely isolated, maintaining both performance and correctness across irregular dimensions.

\subsection{Tiling (stage 3)}
\label{sec:tiling}

After handling the edge cases and splitting the convolution domain accordingly, the next step is to apply loop tilings to the resulting regular kernels.

The core tiling transformation in this stage~\circled{3} is represented in Figure~\ref{fig:flow} by tasks~\circled{3a} and~\circled{3b}.
In~\circled{3a}, the CSA analysis defines the tiling sizes and loop structure according to hardware-aware layout constraints. This first tiling level partitions the kernel iteration space into coarse tiles that match the data layout requirements inferred from the CSA results. In~\circled{3b}, a second tiling level is applied to each coarse tile to match the microkernel (uK) blocking strategy. Here, the iteration space is further subdivided to fit the register-level and cache-level reuse patterns expected by the target architecture.

\begin{figure}[t]
\begin{lstlisting}[language=mlir, caption={Main kernel for \(K_3\) edge case}, label={lst:tiled-kernel}, numbers=left]
%0 = linalg.generic {
  indexing_maps = [
    affine_map<(d0, d1, d2, d3, d4, d5) ->
      (d0, d3, ((d2 floordiv 75) * 2 + d4) * 152 + (d2 mod 75) * 2 + d5)>,
    affine_map<(d0, d1, d2, d3, d4, d5) ->
      (d1, d3, d4, d5)>,
    affine_map<(d0, d1, d2, d3, d4, d5) ->
      (d0, d1, d2)>],
  iterator_types = ["parallel", "parallel", "parallel", "reduction", "reduction", "reduction"]
}
ins(%extracted_slice_3, %extracted_slice_4 : tensor<1x32x22837xf32>, tensor<256x32x3x3xf32>)
outs(%extracted_slice_5 : tensor<1x256x5568xf32>) {
^bb0(%in: f32, %in_15: f32, %out: f32):
  %3 = arith.mulf %in, %in_15 : f32
  %4 = arith.addf %3, %out : f32
  linalg.yield %4 : f32
} -> tensor<1x256x5568xf32>
\end{lstlisting}
\end{figure}

The implementation uses \texttt{scf::tileUsingSCF}, part of the MLIR SCF dialect, to generate structured loop nests for both tiling levels. Immediately after tiling, the loop nest is examined and, if necessary, adjusted~\circled{3c} to enforce the expected loop permutation. This adjustment addresses a specific issue identified during the project, where the MLIR tiling infrastructure consistently fails to propagate the requested \texttt{tilingInterchange} for \new{IS} schedules. The correction process ensures that the innermost loop aligns with the innermost kernel dimension, a requirement both for performance and semantic correctness. A detailed explanation of this issue and its resolution is provided in Section~\ref{sec:lessons}.

\sconvpar{Tiling Strategy for the Main kernel.}
Following edge-case resolution, the main kernel of Listing~\ref{lst:tiled-kernel} example is tiled using a two-level strategy to expose data reuse and enable efficient microkernel execution. Listing~\ref{lst:kernel-tiling} illustrates the transformed loop nest. The first four loops (lines~1--4) correspond to the \textit{outer tiling level}~\circled{3a},  which partitions the convolution along the filters ($K_2$) and spatial output ($K_3$) dimensions. These loops determine the high-level traversal over tiles and coordinate the reuse of packed data for multiple microkernel calls.

In contrast, the two innermost loops (lines~9--10) implement the \textit{inner tiling level}~\circled{3b}. These loops define the iteration space within each tile, operating on register-level blocks of packed input and filter data. Each iteration directly invokes the microkernel, processing a subset of the tile with tight data locality.

This separation between outer and inner tiling enables efficient usage of the memory hierarchy: outer loops control L2/L3 reuse of packed tensors. In contrast, inner loops align with microkernel computation and vectorization strategies.

As discussed in Section~\ref{sec:packing}, packing transformations are applied to prepare the data consumed within the innermost loops. In the example, the input is packed once per tile, and the filters are repacked across the outer tiling level through a process called multipacking.

\begin{figure}[t]
\begin{lstlisting}[language=mlir, caption={Remaining kernel for \(K_3\) edge case (48 elements)} ,label={lst:extra-kernel}, numbers=left]
%1 = linalg.generic {
  indexing_maps = [
    affine_map<(d0, d1, d2, d3, d4, d5) -> 
      (d0, d3, ((d2 floordiv 75) * 2 + d4) * 152 + (d2 mod 75) * 2 + d5)>,
    affine_map<(d0, d1, d2, d3, d4, d5) -> 
      (d1, d3, d4, d5)>,
    affine_map<(d0, d1, d2, d3, d4, d5) -> 
      (d0, d1, d2)>],
  iterator_types = ["parallel", "parallel", "parallel", "reduction", "reduction", "reduction"]
}
ins(%extracted_slice_6, %extracted_slice_7 : tensor<1x32x401xf32>, tensor<256x32x3x3xf32>)
outs(%extracted_slice_8 : tensor<1x256x48xf32>) {
  ^bb0(%in: f32, %in_15: f32, %out: f32):
    %3 = arith.mulf %in, %in_15 : f32
    %4 = arith.addf %3, %out : f32
    linalg.yield %4 : f32
} -> tensor<1x256x48xf32>

\end{lstlisting}
\end{figure}

\begin{figure*}
\begin{lstlisting}[language=mlir, caption={Loop structure for the main kernel after edge-case handling} ,label={lst:kernel-tiling}, numbers=left]
%0 = scf.for %arg3 = 0 to 1 step 1 iter_args(%arg4 = %extracted_slice_5) -> (tensor<1x256x5568xf32>) {
  %3 = scf.for %arg5 = 0 to 32 step 32 iter_args(%arg6 = %arg4) -> (tensor<1x256x5568xf32>) {
    %4 = scf.for %arg7 = 0 to 5568 step 1392 iter_args(%arg8 = %arg6) -> (tensor<1x256x5568xf32>) {
      %5 = scf.for %arg9 = 0 to 256 step 256 iter_args(%arg10 = %arg8) -> (tensor<1x256x5568xf32>) {
        %6 = affine.apply affine_map<(d0) -> (d0 * 2 + (d0 floordiv 75) * 154)>(%arg7)
        %extracted_slice_21 = tensor.extract_slice %extracted_slice_3[%arg3, %arg5, %6] [1, 32, 5861] [1, 1, 1] : tensor<1x32x22837xf32> to tensor<1x32x5861xf32>
        %extracted_slice_22 = tensor.extract_slice %extracted_slice_4[%arg9, %arg5, 0, 0] [256, 32, 3, 3] [1, 1, 1, 1] : tensor<256x32x3x3xf32> to tensor<256x32x3x3xf32>
        %extracted_slice_23 = tensor.extract_slice %arg10[%arg3, %arg9, %arg7] [1, 256, 1392] [1, 1, 1] : tensor<1x256x5568xf32> to tensor<1x256x1392xf32>
        %7 = scf.for %arg11 = 0 to 1392 step 16 iter_args(%arg12 = %extracted_slice_23) -> (tensor<1x256x1392xf32>) {
          %8 = scf.for %arg13 = 0 to 256 step 8 iter_args(%arg14 = %arg12) -> (tensor<1x256x1392xf32>) {
            %9 = affine.apply affine_map<(d0) -> (d0 * 2 + (d0 floordiv 75) * 154)>(%arg11)
            %extracted_slice_29 = tensor.extract_slice %extracted_slice_21[0, 0, %9] [1, 32, 337] [1, 1, 1] : tensor<1x32x5861xf32> to tensor<1x32x337xf32>
            %extracted_slice_30 = tensor.extract_slice %extracted_slice_22[%arg13, 0, 0, 0] [8, 32, 3, 3] [1, 1, 1, 1] : tensor<256x32x3x3xf32> to tensor<8x32x3x3xf32>
            %extracted_slice_31 = tensor.extract_slice %arg14[0, %arg13, %arg11] [1, 8, 16] [1, 1, 1] : tensor<1x256x1392xf32> to tensor<1x8x16xf32>
            %10 = linalg.generic {indexing_maps = [affine_map<(d0, d1, d2, d3, d4, d5) -> (d0, d3, ((d2 floordiv 75) * 2 + d4) * 152 + (d2 mod 75) * 2 + d5)>, affine_map<(d0, d1, d2, d3, d4, d5) -> (d1, d3, d4, d5)>, affine_map<(d0, d1, d2, d3, d4, d5) -> (d0, d1, d2)>], iterator_types = ["parallel", "parallel", "parallel", "reduction", "reduction", "reduction"]} ins(%extracted_slice_29, %extracted_slice_30 : tensor<1x32x337xf32>, tensor<8x32x3x3xf32>) outs(%extracted_slice_31 : tensor<1x8x16xf32>) {
            ^bb0(%in: f32, %in_33: f32, %out: f32):
              %11 = arith.mulf %in, %in_33 : f32
              %12 = arith.addf %11, %out : f32
              linalg.yield %12 : f32
            } -> tensor<1x8x16xf32>
            %inserted_slice_32 = tensor.insert_slice %10 into %arg14[0, %arg13, %arg11] [1, 8, 16] [1, 1, 1] : tensor<1x8x16xf32> into tensor<1x256x1392xf32>
            scf.yield %inserted_slice_32 : tensor<1x256x1392xf32>
          }
          scf.yield %8 : tensor<1x256x1392xf32>
        }
        %inserted_slice_28 = tensor.insert_slice %7 into %arg10[%arg3, %arg9, %arg7] [1, 256, 1392] [1, 1, 1] : tensor<1x256x1392xf32> into tensor<1x256x5568xf32>
        scf.yield %inserted_slice_28 : tensor<1x256x5568xf32>
      }
      scf.yield %5 : tensor<1x256x5568xf32>
    }
    scf.yield %4 : tensor<1x256x5568xf32>
  }
  scf.yield %3 : tensor<1x256x5568xf32>
}
\end{lstlisting}
\end{figure*}

\subsection{Packing (stage 4)}
\label{sec:packing}

Once the loop structure has been corrected, the next stage~\circled{4} transforms the input and filter tensors into packed layouts. This is achieved through affine maps derived from the packing equations detailed in Section~\ref{sec:compat}. The packing process relies on affine computations combined with slice extraction, enabling efficient data reorganization without incurring redundant memory movement. For the input tensor, packing follows a store-in-order scheme, which replicates values across sliding windows to ensure memory alignment and support vectorized access. In contrast, filter packing typically involves only a reordering of elements, with no replication required.

After input and filter packing~\circled{4a}, an optimization step called Multipacking~\circled{4b} is applied to improve data reuse across tiles. This transformation introduces a higher-level grouping along the tiling dimension $K_2$, and is applied selectively based on the execution schedule. In input stationary (IS) scheduling, the filters are multi-packed in groups of $K_2 \times O_c$, enabling reuse of the packed filter tiles across multiple input tiles. Conversely, in the Weight Stationary (WS) schedule, inputs are multipacked in groups of $K_2 \times N_{win}$, enabling reuse of packed input tiles across multiple filter applications. These packing transformations ensure that data are laid out in memory to support efficient access patterns aligned with the microkernel loops, enabling both cache-efficient execution and full vectorization.

After all the transformation stages have been applied, the collapsed spatial dimensions are restored to match the original convolution shape. This Delinearization    task~\circled{4c} ensures compatibility with downstream passes and preserves the expected semantics of the operation.

\sconvpar{Data Packing Strategy for the main kernel.}
Returning to the example in Listing~\ref{lst:kernel-tiling}, where the schedule is \new{IS}, the input Packing task~\circled{4a} is straightforward and occurs once per tile. filters undergo Multipacking~\circled{4b}, \ie{}, are repacked across an outer loop level, grouping $K_2 \times O_c$ elements, to improve cache locality and reuse. These transformations, as shown in Listing~\ref{lst:packing}, convert inefficient strided memory accesses into dense contiguous blocks tailored for locality efficiency.

\begin{figure*}
\begin{lstlisting}[language=mlir, caption={Filter \& Input packing structure for the main kernel} ,label={lst:packing}, numbers=left]
%0 = scf.for %arg3 = 0 to 1 step 1 iter_args(%arg4 = %extracted_slice_5) -> (tensor<1x256x5568xf32>) {
  %3 = scf.for %arg5 = 0 to 32 step 32 iter_args(%arg6 = %arg4) -> (tensor<1x256x5568xf32>) {
    %4 = scf.for %arg7 = 0 to 5568 step 1392 iter_args(%arg8 = %arg6) -> (tensor<1x256x5568xf32>) {
      %5 = scf.for %arg9 = 0 to 256 step 256 iter_args(%arg10 = %arg8) -> (tensor<1x256x5568xf32>) {
        %6 = affine.apply affine_map<(d0) -> ((d0 floordiv 75) * 304 + (d0 mod 75) * 2)>(%arg7)
        %extracted_slice_21 = tensor.extract_slice %extracted_slice_3[%arg3, %arg5, %6] [1, 32, 5861] [1, 1, 1] : tensor<1x32x22837xf32> to tensor<1x32x5861xf32>
        %extracted_slice_22 = tensor.extract_slice %extracted_slice_4[%arg9, %arg5, 0, 0] [256, 32, 3, 3] [1, 1, 1, 1] : tensor<256x32x3x3xf32> to tensor<256x32x3x3xf32>
        %extracted_slice_23 = tensor.extract_slice %arg10[%arg3, %arg9, %arg7] [1, 256, 1392] [1, 1, 1] : tensor<1x256x5568xf32> to tensor<1x256x1392xf32>
        %7 = tensor.empty() : tensor<32x32x3x3x8xf32>
        %8 = linalg.generic {indexing_maps = [affine_map<(d0, d1, d2, d3, d4) -> (d0, d1, d2, d3, d4)>], iterator_types = ["parallel", "parallel", "parallel", "parallel", "parallel"]} outs(%7 : tensor<32x32x3x3x8xf32>) {
        ^bb0(%out: f32):
          %10 = linalg.index 0 : index
          %11 = linalg.index 1 : index
          %12 = linalg.index 2 : index
          %13 = linalg.index 3 : index
          %14 = linalg.index 4 : index
          %15 = affine.apply affine_map<(d0, d1) -> (d0 * 8 + d1)>(%10, %14)
          %extracted = tensor.extract %extracted_slice_22[%15, %11, %12, %13] : tensor<256x32x3x3xf32>
          linalg.yield %extracted : f32
        } -> tensor<32x32x3x3x8xf32>
        %9 = scf.for %arg11 = 0 to 1392 step 16 iter_args(%arg12 = %extracted_slice_23) -> (tensor<1x256x1392xf32>) {
          %10 = affine.apply affine_map<(d0)[s0] -> (((d0 + s0) floordiv 75 - s0 floordiv 75) * 304 + ((d0 + s0) mod 75 - s0 mod 75) * 2)>(%arg11)[%arg7]
          %extracted_slice_29 = tensor.extract_slice %extracted_slice_21[0, 0, %10] [1, 32, 337] [1, 1, 1] : tensor<1x32x5861xf32> to tensor<1x32x337xf32>
          %11 = tensor.empty() : tensor<1x32x3x3x16xf32>
          %12 = linalg.generic {indexing_maps = [affine_map<(d0, d1, d2, d3, d4) -> (d0, d1, d2, d3, d4)>], iterator_types = ["parallel", "parallel", "parallel", "parallel", "parallel"]} outs(%11 : tensor<1x32x3x3x16xf32>) {
          ^bb0(%out: f32):
            %14 = linalg.index 0 : index
            %15 = linalg.index 1 : index
            %16 = linalg.index 2 : index
            %17 = linalg.index 3 : index
            %18 = linalg.index 4 : index
            %19 = affine.apply affine_map<(d0)[s0] -> (d0 + s0)>(%arg11)[%arg7]
            %20 = affine.apply affine_map<(d0, d1, d2)[s0] -> ((((d2 + s0) floordiv 75 - s0 floordiv 75) * 2 + d0) * 152 + ((d2 + s0) mod 75 - s0 mod 75) * 2 + d1)>(%16, %17, %18)[%19]
            %extracted = tensor.extract %extracted_slice_29[%14, %15, %20] : tensor<1x32x337xf32>
            linalg.yield %extracted : f32
          } -> tensor<1x32x3x3x16xf32>
          %collapsed_30 = tensor.collapse_shape %12 [[0], [1, 2, 3], [4]] : tensor<1x32x3x3x16xf32> into tensor<1x288x16xf32>
          %13 = scf.for %arg13 = 0 to 256 step 8 iter_args(%arg14 = %arg12) -> (tensor<1x256x1392xf32>) {
            %14 = affine.apply affine_map<(d0) -> (d0 floordiv 8)>(%arg13)
            %extracted_slice_31 = tensor.extract_slice %8[%14, 0, 0, 0, 0] [1, 32, 3, 3, 8] [1, 1, 1, 1, 1] : tensor<32x32x3x3x8xf32> to tensor<1x32x3x3x8xf32>
            %collapsed_32 = tensor.collapse_shape %extracted_slice_31 [[0, 1, 2, 3], [4]] : tensor<1x32x3x3x8xf32> into tensor<288x8xf32>
            %extracted_slice_33 = tensor.extract_slice %arg14[0, %arg13, %arg11] [1, 8, 16] [1, 1, 1] : tensor<1x256x1392xf32> to tensor<1x8x16xf32>
            %15 = linalg.generic {indexing_maps = [affine_map<(d0, d1, d2, d3) -> (d0, d3, d2)>, affine_map<(d0, d1, d2, d3) -> (d3, d1)>, affine_map<(d0, d1, d2, d3) -> (d0, d1, d2)>], iterator_types = ["parallel", "parallel", "parallel", "reduction"]} ins(%collapsed_30, %collapsed_32 : tensor<1x288x16xf32>, tensor<288x8xf32>) outs(%extracted_slice_33 : tensor<1x8x16xf32>) {
            ^bb0(%in: f32, %in_35: f32, %out: f32):
              %16 = arith.mulf %in, %in_35 : f32
              %17 = arith.addf %16, %out : f32
              linalg.yield %17 : f32
            } -> tensor<1x8x16xf32>
            %inserted_slice_34 = tensor.insert_slice %15 into %arg14[0, %arg13, %arg11] [1, 8, 16] [1, 1, 1] : tensor<1x8x16xf32> into tensor<1x256x1392xf32>
            scf.yield %inserted_slice_34 : tensor<1x256x1392xf32>
          }
          scf.yield %13 : tensor<1x256x1392xf32>
        }
        %inserted_slice_28 = tensor.insert_slice %9 into %arg10[%arg3, %arg9, %arg7] [1, 256, 1392] [1, 1, 1] : tensor<1x256x1392xf32> into tensor<1x256x5568xf32>
        scf.yield %inserted_slice_28 : tensor<1x256x5568xf32>
      }
      scf.yield %5 : tensor<1x256x5568xf32>
    }
    scf.yield %4 : tensor<1x256x5568xf32>
  }
  scf.yield %3 : tensor<1x256x5568xf32>
}


\end{lstlisting}
\end{figure*}

\begin{description}

\item[Filter Multipacking] The multipacking for the filters is defined and allocated early in the transformation (lines~09--20), preparing a dedicated packed tensor. Then a relevant filter slice is extracted and reshaped (lines~40--41) into a 2D packed block with shape $288 \times 8$ (that is, $32 \times 3 \times 3 \times 8$), following the packing equations presented in Section~\ref{sec:packing-affine-modeling}. This layout enables efficient broadcasting and reuse of filters across all output tiles. The packed filter is then consumed directly by the microkernel, as illustrated in line~43.

\item[Input Packing] Similarly, the input tensor is allocated and prepared for packing in lines~24--36. A spatially-windowed slice is extracted and reshaped (line~37) into a packed format with shape $1 \times 288 \times 16$ (that is, $1 \times 32 \times 3 \times 3 \times 16$), which matches the access pattern expected by the microkernel. This step flattens the spatial and kernel dimensions into the inner axes, ensuring alignment with vectorized and cache-friendly computations. As with filters, the packed input is directly used in the microkernel invocation at line~43.

\end{description}

\subsection{MLIR Interface for \SConvOp{}}
\label{sec:api}

\SConvOp{} requires a set of attributes that govern how the convolution transformation is applied. They are architectural features of the target machine and, optionally, user-defined configuration preferences; if not provided, they default to their values. An example of using these attributes is shown in Listing~\ref{lst:sconv-transform}. Such attributes are fed to the Convolution Slicing Analysis (CSA), which then computes all critical tiling and scheduling parameters. CSA takes as input three key descriptors:

\begin{itemize}
  \item \texttt{ConvInfo}, which encodes the structure of the convolution (dimensions of inputs, outputs, and filters);
  \item \texttt{ArchInfo}, which describes the architecture (\eg{}, cache sizes, latencies, and cache line size);
  \item \texttt{mKInfo}, which represents microkernel-specific preferences such as the number of windows (\(N_{win}\)) and filters (\(O_c\)) that the microkernel expects to process per tile (\eg{}, 16 and 8 for Power10 MMA).
\end{itemize}

From these inputs, CSA derives a \texttt{CSAStrategy} object containing:

\begin{itemize}
  \item \texttt{schedule} — the selected schedule (Input Stationary or Weight Stationary);
  \item $N_c$, $K_3$, and $K_2$ — the reduction, and non-reduction tiling sizes aligned with L3 and L2 caches, respectively;
  \item $R_{N_c}$, $R_{K_3}$, and $R_{K_2}$ — values indicating edge cases when the corresponding dimension is not divisible by the selected tile size.
\end{itemize}

The \SConvOp{} allows the user to explicitly override the default values of $N_{win}$, $O_c$, or \texttt{vector\_size}, and to manually set architectural parameters (\texttt{L1\_cache\_size}, \texttt{L2\_cache\_size}, \texttt{L3\_cache\_size}) if automatic detection is not available or desirable. These settings are then propagated into the \texttt{ConvInfo}, \texttt{ArchInfo}, and \texttt{mKInfo} structures passed to CSA.

Importantly, $N_c$, $K_3$, and $K_2$ are not passed as attributes to \SConvOp{} but are computed internally from the full convolution and machine configuration. This separation between declarative configuration and derived strategy ensures that \SConvOp{} remains portable and adaptable to new targets without requiring per-architecture rewrites.

Finally, packing~\circled{4a} and multipacking~\circled{4b} of input and filter tensors are packed according to affine rules for memory alignment and vectorization. This strategy ensures maximum data locality according to the kernel’s reuse pattern.

\begin{figure}[tb]
\begin{lstlisting}[
  language=mlir,
  caption={SConv Transform file.},
  label={lst:sconv-transform},
  numbers=left,
]
module attributes {transform.with_named_sequence} {

  transform.named_sequence @__transform_main(
    %arg0: !transform.any_op) {
    %convs = transform.structured.match ops{["linalg.conv_2d_nchw_fchw"]} in %arg0
      : (!transform.any_op) -> !transform.op<"linalg.conv_2d_nchw_fchw">

    %res_list, %loops_list = transform.structured.sconv %convs
      { mK_info = [16, 8] }
      : (!transform.op<"linalg.conv_2d_nchw_fchw">)
      -> (!transform.op<"linalg.generic">, !transform.any_op)

    transform.yield
  }
}
\end{lstlisting}
\end{figure}

\subsection{Microkernel Lowering (stage 5)}
\label{sec:lowering}

After the Packing stage \circled{4} is finished, the original convolution is transformed into a tiled loop nest that applies packing and calls a microkernel that performs matrix multiplication and accumulation of packed tiles. Such a microkernel is expressed as a \MlirGeneric{} operation for which its inputs and output are of type \texttt{tensor}. The next stage~\circled{5} consists of lowering the microkernel to an optimized code-generated routine or library call. In this work, the BLAS microkernel is used to support that. 

\begin{figure}[tb]
\begin{lstlisting}[
  language=mlir,
  caption={SConv with OpenBLAS calls.},
  label={lst:sconv-transform-blas},
  numbers=left
]
module attributes {transform.with_named_sequence} {

  transform.named_sequence @__transform_main(
    %arg0: !transform.any_op) {

    // SConv Transform (Listing 9)...

    %bufferized = transform.bufferization.one_shot_bufferize %arg0
      { bufferize_function_boundaries = true }
      : (!transform.any_op) -> !transform.any_op

    %funcs = transform.structured.match ops{["func.func"]} in %bufferized
      : (!transform.any_op) -> !transform.any_op

    transform.apply_patterns to %funcs {
      transform.apply_patterns.memref.extract_address_computations
      transform.apply_patterns.memref.expand_strided_metadata
    } : !transform.any_op

    %ukernels_1 = transform.structured.match
      ops{["linalg.generic"]}
      attributes {microkernel} in %bufferized
      : (!transform.any_op) -> !transform.op<"linalg.generic">

    transform.foreach %ukernels_1 : !transform.op<"linalg.generic"> -> !transform.any_op {
    ^bb1(%op: !transform.op<"linalg.generic">):
      %lowered = transform.lower.to_blas "sgemm_blas_kernel", %op
        : (!transform.op<"linalg.generic">) -> (!transform.any_op)
      transform.yield %lowered : !transform.any_op
    }

    transform.yield
  }
}
\end{lstlisting}
\end{figure}

The Transform IR for this stage of the MLIR compilation flow is shown in Listing~\ref{lst:sconv-transform-blas}. It uses all previous transform operations from Listing~\ref{lst:sconv-transform} and the new operations discussed below.

 In step \circled{5a} (Figure~\ref{fig:flow}), bufferization is performed to lower \Tensor{} operations to \Memref{} operations. The \texttt{tensor} type in MLIR is a high-level abstraction whose semantics do not concern memory buffers, allocations, and aliasing. The \texttt{memref} type, on the other hand, is closer to how tensors are represented in numeric libraries: it has a base address, offset, shape, and strides. Bufferization is achieved using the `one-shot bufferize' pass along `extract address computations' and `expand strided metadata' patterns (lines 8-18).

In the next task of this stage \circled{5b} (Figure~\ref{fig:flow}), the microkernel created by the \SConvOp{} must be lowered to an external function call to an optimized microkernel from the OpenBLAS library. High-level BLAS routines, such as SGEMM, perform tiling and packing of their own and call these microkernels, which use available architectural features to accelerate the computation (\eg{}, Intel's AVX512, ARM SME, \del{RISC-V Vector,}and IBM's Power10 MMA). These microkernels are not part of the public API of OpenBLAS; therefore, they are compiled separately and linked to the output of the SConvTransform pipeline.

In concrete terms, a new handle to the microkernel is obtained in the transform IR by matching \MlirGeneric{} operations that have the `microkernel' attribute (lines 20-23). Then, the kernels are iterated over and the \LowerToBlasOp{} is invoked on each of them (lines 25-30). Apart from the microkernel operation handle, this transform operation also expects a string attribute with the name of the function that must be called.

Inside the \LowerToBlasOp{} (lines 27-28), it creates a \MlirExtractStridedMetadata{} that returns the base address, offset, sizes, and strides of the input value. This operation is likely to result in a no-op when lowered to LLVM IR itself, but it is required at this level of abstraction. With the metadata available, the effective address is computed by adding the base address to the offset multiplied by the element width. This is repeated once for the input tile, filter tile, and output tile. With the addresses computed, a \MlirCall{} to the optimized function is created that replaces the generic kernel. The microkernel takes three pointers as parameters that correspond to the input tile, filter tile, and output tile. It also takes as parameters the size and stride of the tiles as integers. All these values can be obtained from the metadata.

Finally, in \circled{5c}, the resulting payload is lowered to the LLVMIR dialect so that is can be executed with \MlirRunner{}. The details for this process are provided in Section~\ref{sec:setup}.

\subsection{Transform Dialect Limitations}
\label{sec:lessons}

The development of SConv as an extension of the MLIR Transform dialect surfaced several limitations and challenges within the existing transformation stack. In addressing these issues, this work implemented specific engineering solutions that may help future enhancements to the Transform dialect.

One limitation encountered involved loop interchange. While \texttt{scf::tileUsingSCF(rewriter, TilingInterfaceOp, TilingOptions)} supports reordering outer loops via the \texttt{setInterchange} method in \texttt{TilingOptions}, this functionality is not applied correctly to the inner tiling levels when scheduling is Input Stationary (IS). To address this, we introduced a custom transformation, \texttt{SwapInductionVars}, which explicitly rewires induction variables and loop bodies to achieve the desired loop ordering in nested tiling. This limitation points to the need for more granular control over loop interchange in the Transform dialect.

Following the tiling transformations, it was observed that adjustments were needed to the affine maps responsible for indexing the linearized input and output tensors. In both regular and rectangular convolutions, the transformations produced index expressions that disrupted the intended memory layout. This led to two distinct levels of adjustment. The first, illustrated in Listings~\ref{lst:kernel-tiling} and~\ref{lst:packing} (lines 5 and 11 in Listing~\ref{lst:kernel-tiling}, and lines 5 and 22 in Listing~\ref{lst:packing}), involved rewriting affine expressions to semantically reconstruct two-dimensional indices from the collapsed spatial dimension properly. The fixed version introduces explicit offset propagation and relative indexing to reconstruct spatial access patterns from a linearized domain.

The second level of adjustment, required in the remaining kernels, addressed more complex scenarios where offsets introduced by prior slicing had to be explicitly integrated into the affine computations. These adjustments ensured consistency and alignment with the original layout assumptions. A representative excerpt illustrating the adjustment applied to the remaining kernel of the example is shown in Listing~\ref{lst:excerpt}, where offset propagation and relative indexing are explicitly encoded to preserve spatial semantics.

\begin{figure}[t]
\begin{lstlisting}[
    language=mlir, 
    caption={Affine map adjustment in the remaining kernel after tiling}, 
    label={lst:excerpt}, 
    numbers=left
]
// after tiling:
  ...
  %7 = scf.for %arg11 = 0 to 48 step 16 iter_args(%arg12 = %extracted_slice_34) -> (tensor<1x256x48xf32>) {
    %8 = scf.for %arg13 = 0 to 256 step 8 iter_args(%arg14 = %arg12) -> (tensor<1x256x48xf32>) {
      %9 = affine.apply affine_map<(d0) -> (d0 * 2 + (d0 floordiv 75) * 154)>(%arg11)

// after corrections:
  ...
  %7 = scf.for %arg11 = 0 to 48 step 16 iter_args(%arg12 = %extracted_slice_34) -> (tensor<1x256x48xf32>) {
    %8 = affine.apply affine_map<(d0) -> (d0 + 5568)>(%arg7)
      %9 = affine.apply affine_map<(d0)[s0] -> (d0 + s0 + 5568)>(%arg11)[%arg7]
      %10 = affine.apply affine_map<()[s0, s1] -> ((s0 floordiv 75 - s1 floordiv 75) * 304 + (s0 mod 75 - s1 mod 75) * 2)>()[%9, %8]
      ...
      %8 = scf.for %arg13 = 0 to 256 step 8 iter_args(%arg14 = %arg12) -> (tensor<1x256x48xf32>) {...
\end{lstlisting}
\end{figure}

Developing the entire transformation pipeline within the Transform dialect, while conceptually modular, required deep familiarity with affine semantics, loop canonicalization, and operand ownership. These complexities are not always abstracted at the Transform dialect level, particularly when manipulating affine maps and composing \MlirExtractSlice{} or \MlirInsertSlice{} operations, indicating an opportunity for improved developer tooling or abstractions.

Furthermore, composing transformation operations programmatically within the \texttt{apply} method of another transformation revealed inconsistencies in reusability. Some transformations, such as \MlirSplit{}, expose reusable interfaces that can be easily invoked, whereas others, like \MlirGeneralize{}, lack the flexibility needed for integration and require partial reimplementation for convolution generalization. Although \textit{SConv} is not simply a composition of existing transformations, many of its components conceptually align with operations already available in the Transform dialect, as discussed throughout this section.

Despite these challenges, the modularity and declarative design of the Transform dialect enabled the construction of a reusable, analyzable, and extensible transformation pipeline. This structure lays the groundwork for future enhancements in convolution code generation and broader compiler infrastructure improvements.

\section{Experimental Evaluation}
\label{sec:experiments}

\begin{table*}[t]
    \centering
    \begin{tabular}{rlll} \toprule 
        CPU info                    & Apple M4        & Intel I7-11700K         & IBM Power E1080\textsuperscript{\dag} \\ \midrule 
        CPU Family (ISA)            & ARM (ARMv9.4-A) & Rocket-Lake (x86-64)    & IBM Power10 (PPC64Ie)   \\ 
        Matrix/Vector Extension     & SME/SVE         & None/AVX512             & MMA/VSX       \\ 
        L1 cache size (KiB)         &  192            & 48                      & 32            \\ 
        L2 cache size (KiB)         &  16384 shared   & 512                     & 1024          \\ 
        L3 cache size (MiB)         &  --             & 16 shared               & 4             \\ 
        CPU frequency (GHz)         & 4.4             & 3.6                     & 2.7           \\ 
        Matrix/Vector Units         & 2/2             & 0/1                     & 1/1           \\ 
        Throughput (GFLOPS/s)       & 502             & 115                     & --            \\ \bottomrule 
    \end{tabular}
    \caption{Summary of evaluated systems. Apple M4 and Intel architectures refer to real evaluations, while IBM Power10 refers to a KVM virtual machine instance. When available, throughput metrics were obtained through microbenchmarks as in \cite{helloSME}. \textsuperscript{\dag}~Power10 host machine specifications.}
    \label{tab:target-achitectures}
\end{table*}

Roughly speaking, the \SConvTransform{} expects a payload file containing \MlirConv{} operations. It will then transform each operation into a tiled loop wrapped around a microkernel call. This section experimentally evaluates two main characteristics of \SConvTransform{}: its completeness and generability. The completeness experiments aim to assess whether \SConvTransform{} generates correct results for different convolution configurations. For this, \SConvTransform{} was evaluated on 7922 convolutions from the Convolution Benchmark (ConvBench~\cite{convbench}) and five different CNN models. The generability experiments, on the other hand, are expected to evaluate whether \SConvTransform{} works and can leverage the Single Instruction Multiple Data (SIMD) capabilities of different architectures. Experiments were performed on architectures that have matrix and vector ISA extensions: (a) two real-world machines (Apple M4 ARM System-on-Chip and Intel 11th-generation i7 processor); and (b) a virtual IBM Power10 machine. 

At first, this section describes the experimental setup that was used. Then, the completeness and generability experiments are discussed, and their results are presented and analyzed.

\subsection{Experimental Setup}
\label{sec:setup}

\SConvTransform{} receives as input a payload file containing \Linalg{} operations, only transforming \MlirConv{} and ignoring all other operations. In order to evaluate the proposed transform, two sets of tests were envisaged. One that takes as input payloads with individual convolutions, and another that takes full CNN model payloads. Each of these payloads is transformed by \SConvTransform{}, generating a tiled and packed \Linalg{} code optimized for convolution.

\subsubsection{Testing set construction:} 
Payloads containing individual convolutions were created based on the ConvBench~\cite{convbench}. ConvBench is a benchmark for evaluating and comparing convolution algorithms in a comprehensive set of 10858 convolution instances acquired from 1280 different DL models\footnote{Numbers were re-acquired with ConvBench provided scripts on May 30, 2025.}. However, grouped convolutions are not subject to \SConvTransform{}; they have their own specialized \Linalg{} operation (\MlirConvGroup{}) that is not suitable for SConv. For this reason, they were filtered out from ConvBench. This resulted in 7922 convolutions, from which 6391 are pointwise ($1 \times 1$ filters), 31 have non-squared filters, and the remaining 1500 refer to regular operations. MLIR payloads were created for all individual convolutions, following a fixed template that wraps the \MlirConv{} convolution operation into an MLIR function. 

The payloads referring to CNN models were exported from the PyTorch DL framework~\cite{pytorch} using the Torch-MLIR project~\cite{torch-mlir}. It first converts the PyTorch model definition into the internal MLIR Torch dialect, and, sequentially, to the target MLIR \Linalg{} dialect. The exported model is encoded as an MLIR function that receives an input tensor and computes the output tensor based on the model computation and saved weights.

For all payloads, an entry-point \texttt{main} function was provided that: (a) creates and initializes the input, weight, and output tensors; (b) uses a \MlirFor{} to repeat 30 times the call of the convolution or the model function to accumulate the execution time; and (c) saves the output tensor elements and computes the average GFLOPS/s performance metric of the convolution based on (b).

\subsubsection{The execution environment:}
 
To execute MLIR code, it must be lowered to the LLVMIR and then proceed to code generation. The \MlirOpt{} utility is used to lower payloads down to the LLVMIR utilizing a set of available conversions and optimization passes. In the remainder of this section, the process of lowering a \Linalg{}-level payload down to the LLVMIR using the ordered passes of Listing~\ref{lst:mlir-opt} is named as \textit{Native Lowering} (stage~\circled{5c} of \figurename~\ref{fig:flow}).

In general, it is possible to directly apply the Native \del{MLIR}Lowering \del{transformations} to the payloads of the testing set. Still, it will naively convert \Linalg{} operations into nested scalar loops without any optimization. For this reason, a common way to lower such operations is to utilize available transformations of the Transform dialect to optimize the operations before lowering to LLVMIR. The proposed \SConvTransform{} is encoded as one of these transformations, which optimizes payloads in the \Linalg{} dialect. It substitutes all encountered \MlirConv{} operations with a tiled loop nest containing packing operations and a microkernel. This initial transformation pass is carried out by the MLIR \TransformOpt{} utility before the Native Lowering. It receives the MLIR payload and the \SConvTransform{} transformation file (Listing~\ref{lst:sconv-transform} of Section~\ref{sec:api}) and returns the optimized \Linalg{} payload.

\begin{figure*}[t]
    \centering
    \includegraphics[width=.9\linewidth]{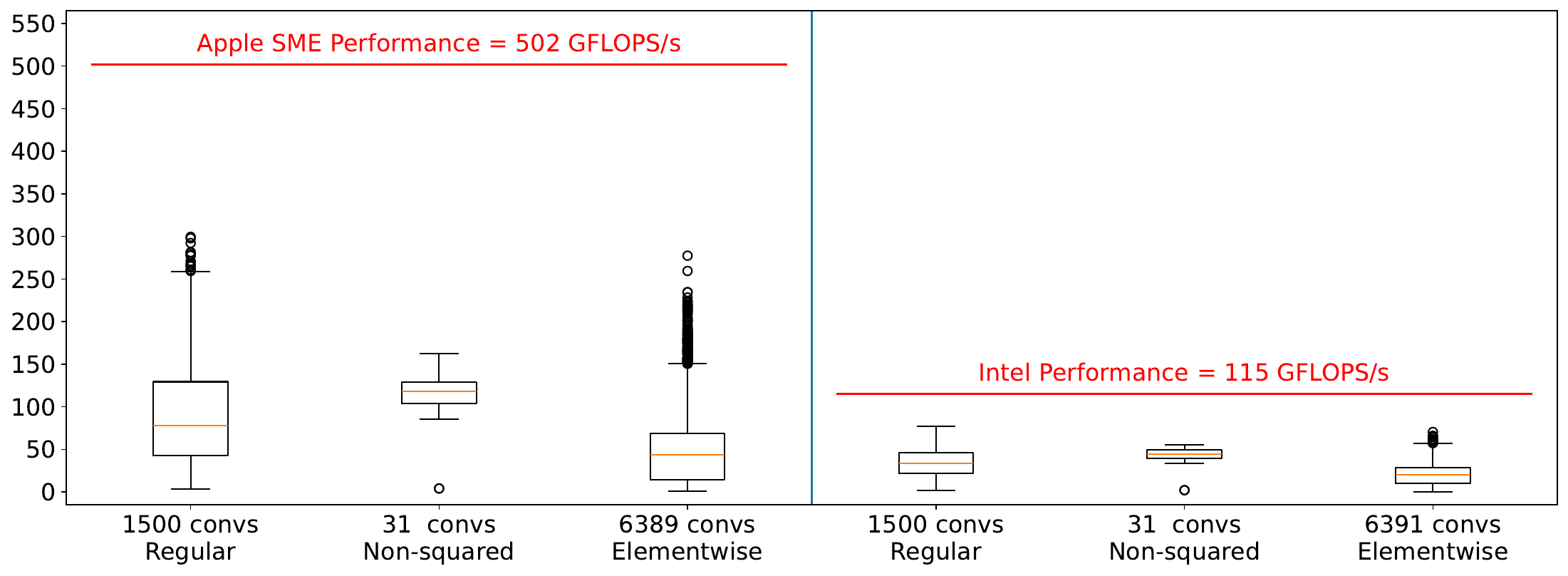}
    \caption{Performance metrics for Apple M4 processor with SME and Intel i7-11700K processor with AVX512.}
    \label{fig:performance}
\end{figure*}

During lowering, the \MlirGeneric{} microkernel added by \SConvTransform{} is still going to be lowered as naive scalar loops by the Native Lowering. One option would be converting the microkernel to \MlirContract{} operations from the Vector dialect and lowering to the target architecture dialect, such as the ArmSME dialect for ARM processors. However, not all architectures have been ported to MLIR, \eg, Power10 does not have consolidated vectorized dialects yet. For this reason, this work proposes substituting the generated microkernel with procedure calls to the super-optimized single precision GEMM (sgemm) OpenBLAS kernels. As described in Section~\ref{sec:lowering}, the \SConvTransform{} file was updated with new passes, \ie, bufferization (stage~\circled{5a}) and the insertion of OpenBLAS microkernel procedure calls (stage~\circled{5b}). After that, the transformed payload with \SConvTransform{} + OpenBLAS microkernel calls is further lowered to the LLVMIR using the Native Lowering.

Finally, the resulting payloads in LLVMIR are Just-in-time executed through the \MlirRunner{} utility. 

\subsection{Completeness Analysis}

The completeness evaluation aims to assess whether the \SConvTransform{} is applicable and produces correct results for all 7922 operations from ConvBench and the five different CNN models.

To evaluate that, each payload is lowered in two different ways. Firstly, it is lowered by the Native Lowering scheme. Then the \SConvTransform{} + OpenBLAS transformation is applied to the payload before also following the Native Lowering. The former generates the baseline execution time and produces the correct outputs, which will be compared with the results produced by the \SConvTransform{} + OpenBLAS lowering. The MLIR files resulting from both lowering paths were then executed by the \MlirRunner{} Just-in-Time (JIT) execution utility on Intel machines, and their results compared. The results revealed that \SConvTransform{} + OpenBLAS generated correct results for all 7922 convolution instances.

In sequence, the LeNet~\cite{lenet}, AlexNet~\cite{alexnet}, VGG19~\cite{vgg19}, SqueezeNet~\cite{squeezenet}, and ConvNext~\cite{convnext} CNN models were evaluated using the \SConvTransform{} + OpenBLAS transformation. They followed the same lowering scheme of ConvBench evaluation, \ie, it first is Native Lowered and executed to obtain the execution time and reference results. Then, it is optimized by the \SConvTransform{} + OpenBLAS transform using the \TransformOpt{} before also being lowered to the LLVMIR using the Native Lowering scheme with the \MlirOpt{} utility. Finally, in either case, the resulting output in LLVMIR is executed using the \MlirRunner{} utility on Intel machines.

\begin{figure}[tb]
\begin{lstlisting}[
  language=sh,
  caption={mlir-opt optimization options.},
  label={lst:mlir-opt},
  numbers=left
]
$ mlir-opt <payload.mlir> \
  --one-shot-bufferize='bufferize-function-boundaries'
  --convert-linalg-to-affine-loops 
  --canonicalize --cse 
  --convert-vector-to-llvm=<target arch> 
  --expand-strided-metadata 
  --lower-affine 
  --convert-scf-to-cf 
  --normalize-memrefs 
  --memref-expand 
  --finalize-memref-to-llvm 
  --lower-affine 
  --convert-func-to-llvm 
  --convert-arith-to-llvm 
  --convert-cf-to-llvm 
  --canonicalize --cse --symbol-dce 
  --llvm-legalize-for-export
\end{lstlisting}
\end{figure}

During the above-described experiment, the Native Lowering step of SqueezeNet and ConvNext CNNs presented some errors, which have been \del{thoroughly} analyzed and fixed as discussed below:

\begin{itemize}
    \item The exported SqueezeNet model contains some \MlirConcat{} operations, which concatenate tensors along a specified dimension. The \MlirOpt{} utility lacks a conversion pattern for this operation. Instead, the \Transform{} dialect provides the \texttt{decompose\_concat} pattern. This transform was added at the beginning of the \SConvTransform{} file (Listing~\ref{lst:sconv-transform}), replacing each \MlirConcat{} with a \MlirEmpty{} of the target shape, followed by a series of \MlirInsertSlice{} operations for each input tensor.
    \item The ConvNext exported payload uses the error function (\MlirErf{}) and reciprocal square-root (\MlirRsqrt{}) from the MLIR \Math{} dialect. To lower Math operations, the \MlirOpt{} tool provides two main conversion passes: \texttt{convert-math-to-llvm}, which lowers \texttt{rsqrt} but not \texttt{erf}; and \texttt{convert-math-to-libm}, which lowers \texttt{erf} but has no mapping for \texttt{rsqrt}. To handle both cases, \texttt{convert-math-to-llvm} is applied first to lower what it can, and \texttt{convert-math-to-libm} is used afterward to handle the remaining ops.
\end{itemize}

After fixing these issues, all model payloads presented correct outputs compared with the baseline.

\subsection{Generability Evaluation}

This section aims to evaluate how agnostic the \SConvTransform{} is concerning the target architecture. As such, it was evaluated on three different machines:
\begin{itemize}
    \item Apple M4 System-on-Chip containing the Scalable Matrix Extension (SME) matrix extension (ARMv9 ISA).
    \item Intel i7-11700K processor with AVX512 vector extension (x86-64 ISA).
    \item IBM Power10 processor running on a KVM virtual machine with Matrix Multiply Assist (MMA) matrix extension.
\end{itemize}
All 7922 convolutions from ConvBench were re-executed on each platform to verify their correctness. 

The platforms have different SIMD hardware capabilities. Apple M4 and IBM Power10 have matrix multiply units, while Intel only has vector units. Furthermore, Apple M4 adopts scalable register sizes, where the implementation dictates its vector/matrix length. On the other hand, Intel and IBM Power10 adopt vector/matrix extensions with lengths predetermined by the instruction set, containing, respectively, 512-bit and 128-bit registers. Given these particularities, the $N_{win}$ and $N_f$ CSA parameters (Section~\ref{sec:sconv} were specially adjusted for the platform's OpenBLAS microkernel implementation. Further platform-dependent information for these architectures is shown in Table~\ref{tab:target-achitectures}.

The payloads were optimized using \SConvTransform{} + OpenBLAS on \TransformOpt{} MLIR program, and the resulting MLIR file was executed on \MlirRunner{} JIT execution engine.

The execution of \SConvTransform{} on these architectures was smooth and presented correct results for all tested models, showing that the approach proposed in this paper is quite generic.

\subsection{Preliminary Performance Analysis}

Please note to the reader that the goal of this work is to evaluate the modularity and generality of the Transform dialect as a way to design a complex convolution operator. It is not the goal of this work to create a highly performant transformation. This will come with time as the work on \SConvTransform{} evolves. That said, a set of experiments was designed to evaluate this transformation performance. 

Additionally, it is not fair to compare the baseline scalar execution performance of Native Lowering against an OpenBLAS-powered SIMD implementation. For this reason, the performance (in GFLOPS/s) of the \SConvTransform{} + OpenBLAS was compared with the system peak performance for the evaluated architecture. The results are shown in \figurename~\ref{fig:performance} for Apple M4 and Intel platforms.

Peak performance was experimentally measured for each real-world target platform by measuring the time of executing 1 billion times a set of the main computation assembly instructions of the OpenBLAS kernel. For Intel, a sequence of 30 \texttt{vfmadd231ps} AVX512 instructions was used to assess peak performance. Each instruction computes two operations (multiply and add) for each of the 16 elements of the vector, resulting in 960 operations per iteration of the billion-step loop. On average, Intel took 8.3 seconds to execute 960 billion operations, resulting in 115 GFLOPS/s. A similar approach was used to calculate the peak performance of the Apple M4 platform. In this case, 32 \texttt{fmopa} SME instructions were organized to accumulate into the \texttt{za0} tile without instruction dependence on the billion-step loop. Each instruction computes $16 \times 16$ multiply and add operations, resulting in 16384 operations per iteration of the billion-step loop. Apple M4 took, on average, approximately $32.5$ seconds to compute $16384$ billion operations, resulting in 502 GFLOPS/s. 

From the experiments, it was clear that the results heavily depend on the convolution configuration. On Apple M4, the median performance for regular convolutions reaches 15.5\% of its peak, going up to 59.6\% in the best case. Intel shows similar behavior, with a median of 28.7\% and a best case of 67\% of its peak. These results can be attributed to several factors. First, peak performance does not reflect real-world behavior due to effects like state flushing from branch mispredictions, instruction dependencies, and memory hierarchy latencies. Additionally, although the OpenBLAS microkernel is expected to be highly optimized, the outer tiling generated by \SConvTransform{} is still unoptimized. It performs repacking and handles edge cases for every macro tile. The outliers with better performance highlight this issue, corresponding to convolutions that reduce repacking or avoid edge cases, thus mitigating the current macrokernel's lack of optimization.

Unfortunately, the Power10 architecture was not suitable for performance evaluation due to the lack of access to real machines and the reliance on virtual machines to perform the experiments. Nevertheless, it also produced correct results in all benchmarks.

\section{Related Works}
\label{sec:related}

This section presents a literature overview of compile-time optimization approaches for linear algebra operations, ranging from optimizations targeted at specific operations~\cite {fft} to whole DL models~\cite{iree,Golin24,tvm}. 

He~and~Markidis~\cite{fft} propose the \texttt{FFTc 2.0}, an MLIR optimization for Fast Fourier Transform (\textit{FFT}) operations. Their approach is similar to \SConvTransform{}, but for FFT operations. They first extended the \Linalg{} dialect with novel FFT operations and then provided mechanisms to lower such operations to machine code. Three main lowering paths were proposed: (a) through handcrafted microkernels; (b) via bufferization and compiler-based vectorization, and (c) outsourcing to GPU kernels. Similarly, the \SConvTransform{} approach leverages the well-known microkernel implementation available in the OpenBLAS library, with enough modularity to run on three different architectures.

Regarding compile-time end-to-end optimization frameworks for DL models, Golin~\etal~\cite{Golin24} proposed to optimize whole ML programs by employing parallelism, tiling, and packing transformations.
However, the user must supply appropriate tile sizes. The microkernel is lowered to a mix of upstream dialects and a dialect called XSMM. \SConvTransform{}, in contrast to this approach, can automatically derive optimal tiling parameters and scheduling strategy through its CSA step.

The Intermediate Representation Execution Environment (IREE)~\cite{iree} is another end-to-end ML compiler with runtime support that fully leverages the MLIR IR under the hood. It uses importers, such as Torch-MLIR~\cite{torch-mlir}, to convert PyTorch models to upstream dialects like Linalg~and~Tensor, TOSA, or MHLO. IREE’s compiler applies data-flow analysis, scheduling, and hardware-specific lowering via its Hardware Abstraction Layer, finally generating machine code through LLVM backend compiler tools. For convolutions, it can lower \texttt{Conv2D} to Im2col or Winograd implementations and optimize them with its flow analyzer. Given that \SConvTransform{} is implemented as a Transform dialect operation, it should seamlessly integrate with IREE.

In contrast to previous works, the TVM~\cite{tvm} compiler does not use the MLIR IR internally. Instead, it automatically explores a comprehensive search space of fusion, tiling, and packing transformations to optimize tensor programs. These transformations are guided by an ML model that predicts program costs on specific hardware, given parameters such as tiling factor and data layout. Ansor~\cite{ansor} builds on TVM by adopting a similar cost model but implementing an evolutionary search over a hierarchical optimization space. Unfortunately, search-based approaches can take minutes for a single operation and hours for an entire model. In contrast, \SConvTransform{}’s analysis relies on straightforward heuristics, producing near-optimal results in just a few seconds.

The proposed \SConvTransform{} extends these ideas by providing automated, compiler-level optimization for convolution operations with minimal runtime overhead. This is achieved by applying the tiling and scheduling parameters determined by CSA analysis directly to a modular kernel implementation. The modular design enables deployment across multiple architectures by leveraging well-known OpenBLAS microkernel implementations.

\section{Conclusion and Future Works}
\label{sec:conclusion}

This work presents \MlirSConv{}, a novel MLIR Transform Dialect operation that optimizes 2D convolution operations by leveraging MLIR’s composable transformation framework. By collapsing the spatial dimensions and applying a static slicing analysis (CSA), SConv enables both generalized and fine-grained transformations such as tiling, packing, and multipacking, even in the presence of irregular configurations.

A key strength of the proposed approach lies in its robust handling of edge cases, such as input or filter regions smaller than tiling thresholds, through region splitting and conditionally applied transformations. Combined with cost-aware scheduling, SConv selects efficient packing strategies that adapt to memory hierarchy characteristics while preserving semantic correctness. 

Experimental results demonstrate the practicality of the method across diverse convolution workloads, with varying hyperparameters such as size, stride, and dilation. When applied to well-structured convolutions, the approach achieves up to 60\% and 67\% of peak performance on Arm SME and Intel AVX512 architectures, respectively, illustrating its effectiveness.

While current packing strategies are selected by CSA using architecture-specific parameters, tile sizes remain configurable within hardware-aware constraints. For future work, it is planned to extend the infrastructure to support automatic padding of tensors, enabling them to participate in the full tiling and packing pipeline. It is also intended to eliminate microkernel invocation overhead by lowering the generic microkernel entirely within MLIR, first to the \Vector{} dialect and then to architecture-specific intrinsics, thereby improving both performance and backend integration.

To further reduce memory pressure during input packing, future work will also explore vector-based packing strategies~\cite{sconv}. This approach reuses data across adjacent tiles by shifting vector registers and appending only the newly required elements, avoiding the need to reload entire vectors. Lastly, in the current release of this work, edge-case handling occurs before tiling and other transformations, resulting in duplicated packing logic and performance penalties. In the future, a new approach will be used to reduce code duplication through more integrated and transformation-aware handling. These enhancements aim to improve further the efficiency, modularity, and portability of the SConv framework.




\bibliographystyle{ACM-Reference-Format}
\bibliography{references}

@misc{Golin24,
  title         = {Towards a high-performance AI compiler with upstream MLIR},
  author        = {R. Golin and 
                   L. Chelini and 
                   A. Siemieniuk and 
                   K. Madhu and 
                   N. Hasabnis and 
                   H. Pabst and 
                   E. Georganas and 
                   A. Heinecke},
  year          = {2024},
  eprint        = {2404.15204},
  archivePrefix = {arXiv},
  primaryClass  = {cs.PL},
  url           = {https://arxiv.org/abs/2404.15204},
}

@inproceedings{transform,
author = {M. P. L\"{u}cke and 
          O. Zinenko and 
          W. S. Moses and 
          M. Steuwer and 
          A. Cohen},
title = {The MLIR Transform Dialect: Your Compiler Is More Powerful Than You Think},
year = {2025},
doi = {10.1145/3696443.3708922},
booktitle = {CGO'25},
pages = {241–254},
}

@article{sconv,
author = {V. Ferrari and 
    R. Sousa and 
    M. Pereira and 
    J. P. L. De Carvalho and 
    J. N. Amaral and 
    J. Moreira and 
    G. Araujo},
title = {Advancing Direct Convolution Using Convolution Slicing Optimization and ISA Extensions},
year = {2023},
volume = {20},
number = {4},
issn = {1544-3566},
doi = {10.1145/3625004},
journal = {ACM Trans. Archit. Code Optim.},
articleno = {54},
numpages = {26},
}

@inproceedings{im2col,
  TITLE = {{High Performance Convolutional Neural Networks for Document Processing}},
  AUTHOR = {K. Chellapilla and 
            S. Puri and 
            P. Simard},
  BOOKTITLE = {{IWFHR'06}},
  YEAR = {2006},
}

@INPROCEEDINGS {winograd,
author = { A. Lavin and 
           S. Gray},
booktitle = {CVPR'16},
title = {{ Fast Algorithms for Convolutional Neural Networks }},
year = {2016},
pages = {4013-4021},
doi = {10.1109/CVPR.2016.435},
}

@inproceedings{convbench,
title={ConvBench: A Comprehensive Benchmark for 2D Convolution Primitive Evaluation},
author={L. Alvarenga and 
        V. Ferrari and 
        R. Sousa and 
        M. Pereira and 
        G. Araujo},
booktitle={MLArchSys'2024},
year={2024},
url={https://openreview.net/forum?id=hSl2W9Zviz}
}

@inproceedings{helloSME,
author = {S. Remke and 
          A. Breuer},
title = {Hello SME! Generating Fast Matrix Multiplication Kernels Using the Scalable Matrix Extension},
year = {2025},
doi = {10.1109/SCW63240.2024.00185},
booktitle = {SC'24},
pages = {1443–1454},
}

@ARTICLE{lenet,
  author={Y. Lecun and 
          L. Bottou and 
          Y. Bengio and 
          P. Haffner},
  journal={Proceedings of the IEEE}, 
  title={Gradient-based learning applied to document recognition}, 
  year={1998},
  volume={86},
  number={11},
  pages={2278-2324},
  doi={10.1109/5.726791}}

@inproceedings{alexnet,
 author = {A. Krizhevsky and 
           I. Sutskever and 
           G. E. Hinton},
 booktitle = {NIPS'12},
 title = {ImageNet Classification with Deep Convolutional Neural Networks},
 year = {2012}
}

@inproceedings{vgg19,
  author       = {K. Simonyan and
                  A. Zisserman},
  title        = {Very Deep Convolutional Networks for Large-Scale Image Recognition},
  booktitle    = {{ICLR}'15},
  year         = {2015},
}

@article{squeezenet,
  author       = {F. N. Iandola and
                  M. W. Moskewicz and
                  K. Ashraf and
                  S. Han and
                  W. J. Dally and
                  K. Keutzer},
  title        = {SqueezeNet: AlexNet-level accuracy with 50x fewer parameters and {\textless}1MB
                  model size},
  journal      = {CoRR},
  volume       = {abs/1602.07360},
  year         = {2016},
}

@inproceedings{convnext,
  author       = {Z. Liu and
                  H. Mao and
                  C{-}Y. Wu and
                  C. Feichtenhofer and
                  T. Darrell and
                  S. Xie},
  title        = {A ConvNet for the 2020s},
  booktitle    = {CVPR'22},
  pages        = {11966--11976},
  year         = {2022},
  doi          = {10.1109/CVPR52688.2022.01167},
}

@article{MMA,
  author    = {J. E. Moreira and
               K. Barton and
               S. Battle and
               P. Bergner and
               R. Bertran and
               P. Bhat and
               P. Caldeira and
               D. Edelsohn and
               G. Fossum and
               B. Frey and
               N. Ivanovic and
               C. Kerchner and
               V. Lim and
               S. Kapoor and
               T. M. Filho and
               S. M. Mueller and
               B. Olsson and
               S. Sadasivam and
               B. Saleil and
               B. Schmidt and
               R. Srinivasaraghavan and
               S. Srivatsan and
               B. W. Thompto and
               A. Wagner and
               N. Wu},
  title     = {A matrix math facility for Power {ISA(TM)} processors},
  journal   = {CoRR},
  volume    = {abs/2104.03142},
  year      = {2021},
}

@inproceedings{tvm,
  author    = {T. Chen and 
               T. Moreau and 
               Z. Jiang and 
               L. Zheng and 
               E. Yan and 
               H. Shen and 
               M. Cowan and 
               L. Wang and 
               Y. Hu and 
               L. Ceze and 
               C. Guestrin and 
               A. Krishnamurthy},
  title     = {{TVM}: An Automated {End-to-End} Optimizing Compiler for Deep Learning},
  booktitle = {OSDI'18},
  year      = {2018},
  pages     = {578--594},
}

@inproceedings {ansor,
    author    = {L. Zheng and 
                 C. Jia and 
                 M. Sun and 
                 Z. Wu and 
                 C. Hao Yu and 
                 A. Haj-Ali and 
                 Y. Wang and 
                 J. Yang and 
                 D. Zhuo and 
                 K. Sen and 
                 J. E. Gonzalez and 
                 I. Stoica},
    title     = {Ansor: Generating {High-Performance} Tensor Programs for Deep Learning},
    booktitle = {OSDI'20},
    year      = {2020},
    pages     = {863--879},
}

@inproceedings{pytorch,
    author = {J. Ansel and 
              E. Yang and 
              H. He and 
              N. Gimelshein and 
              A. Jain and 
              M. Voznesensky and 
              B. Bao and 
              P. Bell and 
              D. Berard and 
              E. Burovski and 
              G. Chauhan and 
              A. Chourdia and 
              W. Constable and 
              A. Desmaison and 
              Z. DeVito and 
              E. Ellison and 
              W. Feng and 
              J. Gong and 
              M. Gschwind and 
              B. Hirsh and 
              S. Huang and 
              K. Kalambarkar and 
              L. Kirsch and 
              M. Lazos and 
              M. Lezcano and 
              Y. Liang and 
              J. Liang and 
              Y. Lu and 
              C-K. Luk and 
              B. Maher and 
              Y. Pan and 
              C. Puhrsch and 
              M. Reso and 
              M. Saroufim and 
              M. Y. Siraichi and 
              H. Suk and 
              M. Suo and 
              P. Tillet and 
              E. Wang and 
              X. Wang and 
              W. Wen and 
              S. Zhang and 
              X. Zhao and 
              K. Zhou and 
              R. Zou and 
              A. Mathews and 
              G. Chanan and 
              P. Wu and 
              S. Chintala},
    booktitle = {ASPLOS'24},
    doi = {10.1145/3620665.3640366},
    title = {{PyTorch 2: Faster Machine Learning Through Dynamic Python Bytecode Transformation and Graph Compilation}},
    pages = {929--947},
    year = {2024}
}

@inproceedings{fft,
  author       = {Y. He and
                  S. Markidis},
  title        = {High-Performance {FFT} Code Generation via {MLIR} Linalg Dialect and
                  {SIMD} Micro-Kernels},
  booktitle    = {{CLUSTER}'2024},
  pages        = {155--165},
  year         = {2024},
  doi          = {10.1109/CLUSTER59578.2024.00021}
}

@inproceedings{mlir,
  author       = {C. Lattner and
                  M. Amini and
                  U. Bondhugula and
                  A. Cohen and
                  A. Davis and
                  J. A. Pienaar and
                  R. Riddle and
                  T. Shpeisman and
                  N. Vasilache and
                  O. Zinenko},
  title        = {{MLIR:} Scaling Compiler Infrastructure for Domain Specific Computation},
  booktitle    = {CGO'2021},
  pages        = {2--14},
  year         = {2021},
  doi          = {10.1109/CGO51591.2021.9370308}
}

@software{torch-mlir,
    author = {{LLVM}},
    license = {["Apache-2.0 with LLVM Exceptions", "BSD"]},
    title = {{Torch-MLIR}},
    url = {https://github.com/llvm/torch-mlir}
}

@software{iree,
author = {{The IREE Authors}},
license = {Apache-2.0 WITH LLVM-exception},
title = {{IREE}},
url = {https://github.com/iree-org/iree},
year = {2019}
}

@ARTICLE{cal_2024_kim,
    author={H. Kim and 
            G. Ye and 
            N. Wang and 
            A. Yazdanbakhsh and 
            N. S. Kim},
    journal={IEEE Computer Architecture Letters},
    title={{ Exploiting Intel Advanced Matrix Extensions (AMX) for Large Language Model Inference }},
    year={2024},
    volume={23},
    number={01},
    ISSN={1556-6064},
    pages={117-120},
    doi={10.1109/LCA.2024.3397747},
}

@manual{intel,
  title        = {Intel\textsuperscript{\textregistered} 64 and IA-32 Architectures Instruction Set Extensions Programming Reference},
  author       = {{Intel Corporation}},
  year         = {2025},
  note         = {Order Number: 319433-058},
  url          = {https://www.intel.com/content/www/us/en/developer/articles/technical/intel-sdm.html}
}

\end{document}